# Adaptive Stochastic Resource Control:
# A Machine Learning Approach

**Balázs Csanád Csáji**                                    BALAZS.CSAJI@SZTAKI.HU
*Computer and Automation Research Institute,*
*Hungarian Academy of Sciences*
*Kende utca 13–17, Budapest, H–1111, Hungary*

**László Monostori**                                    LASZLO.MONOSTORI@SZTAKI.HU
*Computer and Automation Research Institute,*
*Hungarian Academy of Sciences; and*
*Faculty of Mechanical Engineering,*
*Budapest University of Technology and Economics*

## Abstract

The paper investigates stochastic resource allocation problems with scarce, reusable resources and non-preemtive, time-dependent, interconnected tasks. This approach is a natural generalization of several standard resource management problems, such as scheduling and transportation problems. First, reactive solutions are considered and defined as control policies of suitably reformulated Markov decision processes (MDPs). We argue that this reformulation has several favorable properties, such as it has finite state and action spaces, it is aperiodic, hence all policies are proper and the space of control policies can be safely restricted. Next, approximate dynamic programming (ADP) methods, such as fitted Q-learning, are suggested for computing an efficient control policy. In order to compactly maintain the cost-to-go function, two representations are studied: hash tables and support vector regression (SVR), particularly, $\nu$-SVRs. Several additional improvements, such as the application of limited-lookahead rollout algorithms in the initial phases, action space decomposition, task clustering and distributed sampling are investigated, too. Finally, experimental results on both benchmark and industry-related data are presented.

## 1. Introduction

*Resource allocation problems* (RAPs) are of high practical importance, since they arise in many diverse fields, such as manufacturing production control (e.g., production scheduling), warehousing (e.g., storage allocation), fleet management (e.g., freight transportation), personnel management (e.g., in an office), scheduling of computer programs (e.g., in massively parallel GRID systems), managing a construction project or controlling a cellular mobile network. RAPs are also central to management science (Powell & Van Roy, 2004). In the paper we consider optimization problems that include the assignment of a finite set of reusable resources to non-preemtive, interconnected tasks that have stochastic durations and effects. Our objective is to investigate efficient reactive (closed-loop) decision-making processes that can deal with the allocation of scarce resources over time with a goal of optimizing the objectives. For "real world" applications, it is important that the solution should be able to deal with large-scale problems and handle environmental changes, as well.

## 1.1 Industrial Motivations

One of our main motivations for investigating RAPs is to enhance manufacturing production control. Regarding contemporary manufacturing systems, difficulties arise from unexpected tasks and events, non-linearities, and a multitude of interactions while attempting to control various activities in dynamic shop floors. Complexity and uncertainty seriously limit the effectiveness of conventional production control approaches (e.g., deterministic scheduling).

In the paper we apply mathematical programming and *machine learning* (ML) techniques to achieve the suboptimal control of a general class of stochastic RAPs, which can be vital to an *intelligent manufacturing system* (IMS). The term of IMS can be attributed to a tentative forecast of Hatvany and Nemes (1978). In the early 80s IMSs were outlined as the next generation of manufacturing systems that utilize the results of artificial intelligence research and were expected to solve, within certain limits, unprecedented, unforeseen problems on the basis of even incomplete and imprecise information. Naturally, the applicability of the different solutions to RAPs are not limited to industrial problems.

## 1.2 Curse(s) of Dimensionality

Different kinds of RAPs have a huge number of exact and approximate solution methods, for example, in the case of scheduling problems (Pinedo, 2002). However, these methods primarily deal with the static (and often strictly deterministic) variants of the various problems and, mostly, they are not aware of uncertainties and changes. Special (deterministic) RAPs which appear in the field of *combinatorial optimization*, such as the traveling salesman problem (TSP) or the job-shop scheduling problem (JSP), are *strongly NP-hard* and, moreover, they do not have any good polynomial-time approximation, either.

In the stochastic case RAPs can be often formulated as *Markov decision processes* (MDPs) and by applying *dynamic programming* (DP) methods, in theory, they can be solved *optimally*. However, due to the phenomenon that was named *curse of dimensionality* by Bellman (1961), these methods are highly intractable in practice. The "curse" refers to the combinatorial explosion of the required computation as the size of the problem increases. Some authors, e.g., Powell and Van Roy (2004), talk about even three types of curses concerning DP algorithms. This has motivated *approximate* approaches that require a more tractable computation, but often yield *suboptimal* solutions (Bertsekas, 2005).

## 1.3 Related Literature

It is beyond our scope to give a general overview on different solutions to RAPs, hence, we only concentrate on the part of the literature that is closely related to our approach. Our solution belongs to the class of *approximate dynamic programming* (ADP) algorithms which constitute a broad class of discrete-time control techniques. Note that ADP methods that take an actor-critic point of view are often called *reinforcement learning* (RL).

Zhang and Dietterich (1995) were the first to apply an RL technique for a special RAP. They used the $TD(\lambda)$ method with iterative repair to solve a static scheduling problem, namely, the NASA space shuttle payload processing problem. Since then, a number of papers have been published that suggested using RL for different RAPs. The first reactive (closed-loop) solution to scheduling problems using ADP algorithms was briefly described by Schneider, Boyan, and Moore (1998). Riedmiller and Riedmiller (1999) used a *multi-layer perceptron* (MLP) based neural RL approach to learn local heuristics. Aydin and Öztemel (2000) applied a modified version of Q-learning to learn dispatching rules for production scheduling. Multi-agent versions of ADP techniques for solving dynamic scheduling problems were also suggested (Csáji, Kádár, & Monostori, 2003; Csáji & Monostori, 2006).

Powell and Van Roy (2004) presented a formal framework for RAPs and they applied ADP to give a general solution to their problem. Later, a parallelized solution to the previously defined problem was given by Topaloglu and Powell (2005). Our RAP framework, presented in Section 2, differs from these approaches, since in our system the goal is to accomplish a set of tasks that can have widely different stochastic durations and precedence constraints between them, while Powell and Van Roy's (2004) approach concerns with satisfying many similar demands arriving stochastically over time with demands having unit durations but not precedence constraints. Recently, *support vector machines* (SVMs) were applied by Gersmann and Hammer (2005) to improve iterative repair (local search) strategies for *resource constrained project scheduling problems* (RCPSPs). An agent-based resource allocation system with MDP-induced preferences was presented by Dolgov and Durfee (2006). Finally, Beck and Wilson (2007) gave proactive solutions for job-shop scheduling problems based on the combination of *Monte Carlo simulation*, solutions of the associated deterministic problem, and either constraint programming or tabu-search.

## 1.4 Main Contributions

As a summary of the main contributions of the paper, it can be highlighted that:

1. We propose a formal framework for investigating stochastic resource allocation problems with scarce, reusable resources and non-preemtive, time-dependent, interconnected tasks. This approach constitutes a natural generalization of several standard resource management problems, such as scheduling problems, transportation problems, inventory management problems or maintenance and repair problems.

   This general RAP is reformulated as a stochastic shortest path problem (a special MDP) having favorable properties, such as, it is aperiodic, its state and action spaces are finite, all policies are proper and the space of control policies can be safely restricted. Reactive solutions are defined as policies of the reformulated problem.

2. In order to compute a good approximation of the optimal control policy, ADP methods are suggested, particularly, fitted Q-learning. Regarding value function representations for ADP, two approaches are studied: hash tables and SVRs. In the latter, the samples for the regression are generated by Monte Carlo simulation and in both cases the inputs are suitably defined numerical feature vectors.

   Several improvements to speed up the calculation of the ADP-based solution are suggested: application of limited lookahead rollout algorithms in the initial phases to guide the exploration and to provide the first samples to the approximator; decomposing the action space to decrease the number of available actions in the states; clustering the tasks to reduce the length of the trajectories and so the variance of the cumulative costs; as well as two methods to distribute the proposed algorithm among several processors having either a shared or a distributed memory architecture.

3. The paper also presents several results of numerical experiments on both benchmark and industry-related problems. First, the performance of the algorithm is measured on hard benchmark flexible job-shop scheduling datasets. The scaling properties of the approach are demonstrated by experiments on a simulated factory producing mass-products. The effects of clustering depending on the size of the clusters and the speedup relative to the number of processors in case of distributed sampling are studied, as well. Finally, results on the adaptive features of the algorithm in case of disturbances, such as resource breakdowns or new task arrivals, are also shown.

## 2. Markovian Resource Control

This section aims at precisely defining RAPs and reformulating them in a way that would allow them to be effectively solved by ML methods presented in Section 3. First, a brief introduction to RAPs is given followed by the formulation of a general resource allocation framework. We start with deterministic variants and then extend the definition to the stochastic case. Afterwards, we give a short overview on *Markov decision processes* (MDPs), as they constitute a fundamental theory to our approach. Next, we reformulate the reactive control problem of RAPs as a *stochastic shortest path* (SSP) problem (a special MDP).

### 2.1 Classical Problems

In this section we give a brief introduction to RAPs through two strongly NP-hard combinatorial optimization problems: the job-shop scheduling problem and the traveling salesman problem. Later, we will apply these two basic problems to demonstrate the results.

#### 2.1.1 JOB-SHOP SCHEDULING

First, we consider the classical *job-shop scheduling problem* (JSP) which is a standard deterministic RAP (Pinedo, 2002). We have a set of jobs, $\mathcal{J} = \{J_1, \ldots, J_n\}$, to be processed through a set of machines, $\mathcal{M} = \{M_1, \ldots, M_k\}$. Each $j \in \mathcal{J}$ consists of a sequence of $n_j$ tasks, for each task $t_{ji} \in \mathcal{T}$, where $i \in \{1, \ldots, n_j\}$, there is a machine $m_{ji} \in \mathcal{M}$ which can process the task, and a processing time $p_{ji} \in \mathbb{N}$. The aim of the optimization is to find a *feasible schedule* that minimizes a given performance measure. A solution, i.e., a schedule, is a suitable "task to starting time" assignment. Figure 1 visualizes an example schedule by using a Gantt chart. Note that a Gantt chart (Pinedo, 2002) is a figure using bars, in order to illustrate the starting and finishing times of the tasks on the resources.

The concept of "feasibility" will be defined in Section 2.2.1. In the case of JSP a feasible schedule can be associated with an ordering of the tasks, i.e., the order in which they will be executed on the machines. There are many types of performance measures available for JSP, but probably the most commonly applied one is the maximum completion time of the tasks, also called "makespan". In case of applying makespan, JSP can be interpreted as the problem of finding a schedule which completes all tasks in every job as soon as possible.

Later, we will study an extension of JSP, called the *flexible job-shop scheduling problem* (FJSP) that arises when some of the machines are interchangeable, i.e., there may be tasks that can be executed on several machines. In this case the processing times are given by a *partial* function, $p : \mathcal{M} \times \mathcal{T} \hookrightarrow \mathbb{N}$. Note that a partial function is a binary relation that associates the elements of its domain set with *at most* one element of its range set. Throughout the paper we use "↪" to denote partial function type binary relations.

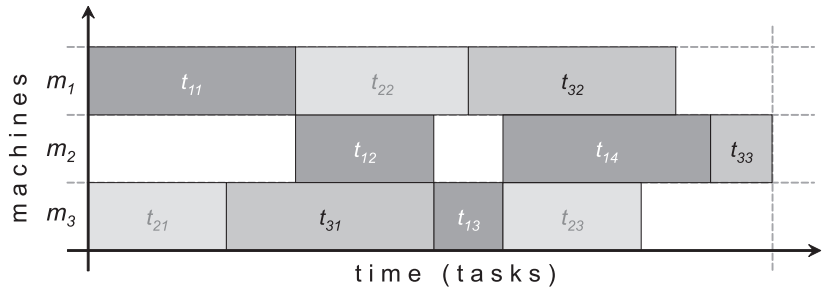

Figure 1: A possible solution to JSP, presented in a Gantt chart. Tasks having the same color belong to the same job and should be processed in the given order. The vertical gray dotted line indicates the maximum completion time of the tasks.

### 2.1.2 TRAVELING SALESMAN

One of the basic logistic problems is the *traveling salesman problem* (TSP) that can be stated as follows. Given a number of cities and the costs of travelings between them, which is the least-cost round-trip route that visits each city exactly once and then returns to the starting city. Several variants of TSP are known, the most standard one can be formally characterized by a connected, undirected, edge-weighted graph $G = \langle V, E, w \rangle$, where $V = \{1, \ldots, n\}$ is the vertex set corresponding to the set of "cities", $E \subseteq V \times V$ is the set of edges which represents the "roads" between the cities, and $w : E \to \mathbb{N}$ defines the weights of the edges: the durations of the trips. The aim of the optimization is to find a *Hamilton-circuit* with the smallest possible weight. Note that a Hamilton-circuit is a graph cycle that starts at a vertex, passes through every vertex exactly once, and returns to the starting vertex. Take a look at Figure 2 for an example Hamilton-circuit.

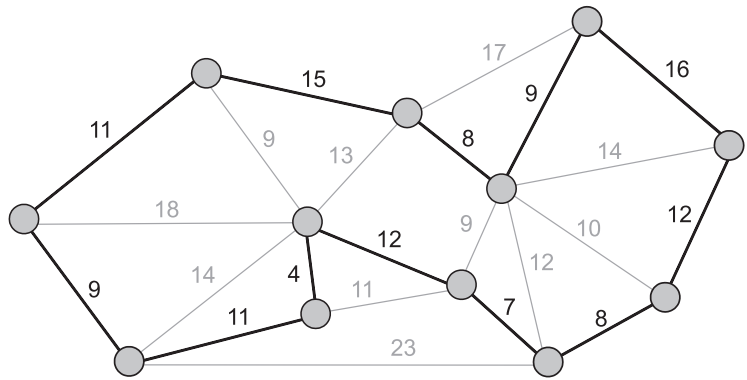

Figure 2: A possible solution to TSP, a closed path in the graph. The black edges constitute a Hamilton-circuit in the given connected, undirected, edge-weighted graph.

## 2.2 Deterministic Framework

Now, we present a general framework to model resource allocation problems. This framework can be treated as a generalization of several classical RAPs, such as JSP and TSP.

First, a deterministic resource allocation problem is considered: an instance of the problem can be characterized by an 8-tuple $\langle \mathcal{R}, \mathcal{S}, \mathcal{O}, \mathcal{T}, \mathcal{C}, d, e, i \rangle$. In details the problem consists of a set of reusable *resources* $\mathcal{R}$ together with $\mathcal{S}$ that corresponds to the set of possible *resource states*. A set of allowed *operations* $\mathcal{O}$ is also given with a subset $\mathcal{T} \subseteq \mathcal{O}$ which denotes the *target operations* or *tasks*. $\mathcal{R}$, $\mathcal{S}$ and $\mathcal{O}$ are supposed to be finite and they are pairwise disjoint. There can be *precedence constrains* between the tasks, which are represented by a partial ordering $\mathcal{C} \subseteq \mathcal{T} \times \mathcal{T}$. The *durations* of the operations depending on the state of the executing resource are defined by a *partial* function $d : \mathcal{S} \times \mathcal{O} \hookrightarrow \mathbb{N}$, where $\mathbb{N}$ is the set of natural numbers, thus, we have a discrete-time model. Every operation can *affect* the state of the executing resource, as well, that is described by $e : \mathcal{S} \times \mathcal{O} \hookrightarrow \mathcal{S}$ which is also a partial function. It is assumed that $dom(d) = dom(e)$, where $dom(\cdot)$ denotes the domain set of a function. Finally, the *initial states* of the resources are given by $i : \mathcal{R} \to \mathcal{S}$.

The state of a resource can contain all relevant information about it, for example, its type and current setup (scheduling problems), its location and load (transportation problems) or condition (maintenance and repair problems). Similarly, an operation can affect the state in many ways, e.g., it can change the setup of the resource, its location or condition. The system must allocate each task (target operation) to a resource, however, there may be cases when first the state of a resource must be modified in order to be able to execute a certain task (e.g., a transporter may need, first, to travel to its loading/source point, a machine may require repair or setup). In these cases non-task operations may be applied. They can modify the states of the resources without directly serving a demand (executing a task). It is possible that during the resource allocation process a *non-task* operation is applied several times, but other *non-task* operations are completely avoided (for example, because of their high cost). Nevertheless, finally, all *tasks* must be completed.

### 2.2.1 FEASIBLE RESOURCE ALLOCATION

A *solution* for a deterministic RAP is a partial function, the *resource allocator function*, $\varrho : \mathcal{R} \times \mathbb{N} \hookrightarrow \mathcal{O}$ that assigns the *starting times* of the operations on the resources. Note that the operations are supposed to be *non-preemptive* (they may not be interrupted).

A solution is called *feasible* if and only if the following four properties are satisfied:

1. All tasks are associated with exactly one (resource, time point) pair:
   $\forall v \in \mathcal{T} : \exists! \langle r, t \rangle \in dom(\varrho) : v = \varrho(r, t)$.

2. Each resource executes, at most, one operation at a time:
   $\neg \exists u, v \in \mathcal{O} : u = \varrho(r, t_1) \land v = \varrho(r, t_2) \land t_1 \leq t_2 < t_1 + d(s(r, t_1), u)$.

3. The precedence constraints on the tasks are satisfied:
   $\forall \langle u, v \rangle \in \mathcal{C} : [u = \varrho(r_1, t_1) \land v = \varrho(r_2, t_2)] \Rightarrow [t_1 + d(s(r_1, t_1), u) \leq t_2]$.

4. Every operation-to-resource assignment is valid:
   $\forall \langle r, t \rangle \in dom(\varrho) : \langle s(r, t), \varrho(r, t) \rangle \in dom(d)$,

where $s : \mathcal{R} \times \mathbb{N} \to \mathcal{S}$ describes the states of the resources at given times

$$s(r, t) = \begin{cases} i(r) & \text{if } t = 0 \\ s(r, t-1) & \text{if } \langle r, t \rangle \notin dom(\varrho) \\ e(s(r, t-1), \varrho(r, t)) & \text{otherwise} \end{cases}$$

A RAP is called *correctly specified* if there exists at least one feasible solution. In what follows it is assumed that the problems are correctly specified. Take a look at Figure 3.

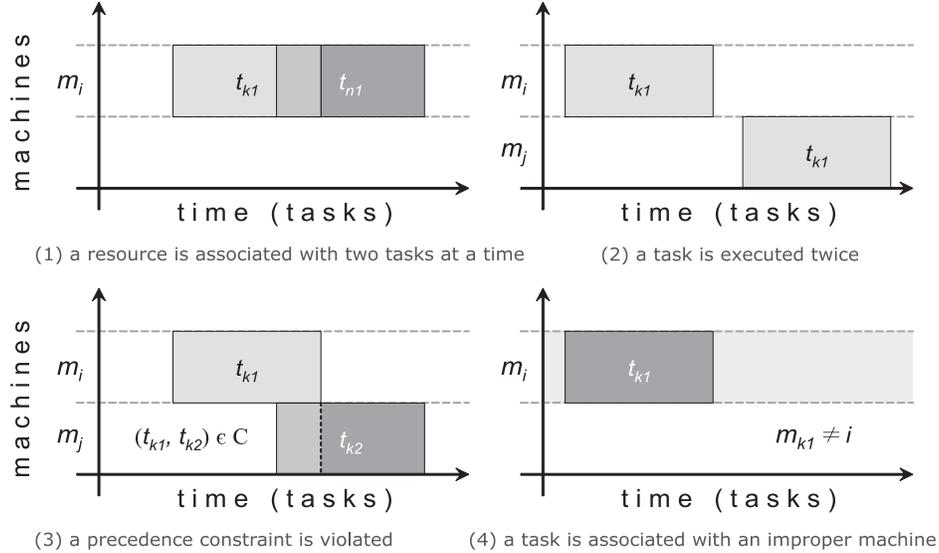

(1) a resource is associated with two tasks at a time

(2) a task is executed twice

(3) a precedence constraint is violated

(4) a task is associated with an improper machine

Figure 3: Feasibility – an illustration of the four forbidden properties, using JSP as an example. The presented four cases are excluded from the set of feasible schedules.

### 2.2.2 Performance Measures

The set of all feasible solutions is denoted by $\mathbb{S}$. There is a performance (or cost) associated with each solution, which is defined by a *performance measure* $\kappa : \mathbb{S} \to \mathbb{R}$ that often depends on the task completion times, only. Typical performance measures that appear in practice include: maximum completion time or mean flow time. The aim of the resource allocator system is to compute a feasible solution with maximal performance (or minimal cost).

Note that the performance measure can assign penalties for violating *release* and *due dates* (if they are available) or can even reflect the *priority* of the tasks. A possible generalization of the given problem is the case when the operations may require more resources simultaneously, which is important to model, e.g., resource constrained project scheduling problems. However, it is straightforward to extend the framework to this case: the definition of $d$ and $e$ should be changed to $d : \mathcal{S}^{\langle k \rangle} \times \mathcal{O} \to \mathbb{N}$ and $e : \mathcal{S}^{\langle k \rangle} \times \mathcal{O} \to \mathcal{S}^{\langle k \rangle}$, where $\mathcal{S}^{\langle k \rangle} = \cup_{i=1}^{k} \mathcal{S}^i$ and $k \leq |\mathcal{R}|$. Naturally, we assume that for all $\langle \hat{s}, o \rangle \in dom(e) : dim(e(\hat{s}, o)) = dim(\hat{s})$. Although, managing tasks with multiple resource requirements may be important in some cases, to keep the analysis as simple as possible, we do not deal with them in the paper. Nevertheless, most of the results can be easily generalized to that case, as well.

### 2.2.3 DEMONSTRATIVE EXAMPLES

Now, as demonstrative examples, we reformulate (F)JSP and TSP in the given framework.

It is straightforward to formulate scheduling problems, such as JSP, in the presented resource allocation framework: the tasks of JSP can be directly associated with the tasks of the framework, machines can be associated with resources and processing times with durations. The precedence constraints are determined by the linear ordering of the tasks in each job. Note that there is only one possible resource state for every machine. Finally, feasible schedules can be associated with feasible solutions. If there were setup-times in the problem, as well, then there would be several states for each resource (according to its current setup) and the "set-up" procedures could be associated with the non-task operations.

A RAP formulation of TSP can be given as follows. The set of resources consists of only one element, namely the "salesman", therefore, $\mathcal{R} = \{r\}$. The possible states of resource $r$ (the salesman) are $\mathcal{S} = \{s_1, \ldots, s_n\}$. If the state (of $r$) is $s_i$, it indicates that the salesman is in city $i$. The allowed operations are the same as the allowed tasks, $\mathcal{O} = \mathcal{T} = \{t_1, \ldots, t_n\}$, where the execution of task $t_i$ symbolizes that the salesman travels to city $i$ from his current location. The constraints $\mathcal{C} = \{\langle t_2, t_1 \rangle, \langle t_3, t_1 \rangle \ldots, \langle t_n, t_1 \rangle\}$ are used for forcing the system to end the whole round-tour in city 1, which is also the starting city, thus, $i(r) = s_1$. For all $s_i \in \mathcal{S}$ and $t_j \in \mathcal{T}$: $\langle s_i, t_j \rangle \in dom(d)$ if and only if $\langle i, j \rangle \in E$. For all $\langle s_i, t_j \rangle \in dom(d): d(s_i, t_j) = w_{ij}$ and $e(s_i, t_j) = s_j$. Note that $dom(e) = dom(d)$ and the first feasibility requirement guarantees that each city is visited exactly once. The performance measure $\kappa$ is the latest arrival time, $\kappa(\varrho) = \max\{t + d(s(r,t), \varrho(r,t)) \mid \langle r, t \rangle \in dom(\varrho)\}$.

### 2.2.4 COMPUTATIONAL COMPLEXITY

If we use a performance measure which has the property that a solution can be precisely defined by a bounded sequence of operations (which includes all tasks) with their assignment to the resources and, additionally, among the solutions generated this way an optimal one can be found, then the RAP becomes a *combinatorial optimization* problem. Each performance measure monotone in the completion times, called *regular*, has this property. Because the above defined RAP is a generalization of, e.g., JSP and TSP, it is *strongly NP-hard* and, furthermore, no good polynomial-time approximation of the optimal resource allocating algorithm exits, either (Papadimitriou, 1994).

## 2.3 Stochastic Framework

So far our model has been deterministic, now we turn to stochastic RAPs. The stochastic variant of the described general class of RAPs can be defined by randomizing functions $d$, $e$ and $i$. Consequently, the operation durations become random, $d : \mathcal{S} \times \mathcal{O} \to \Delta(\mathbb{N})$, where $\Delta(\mathbb{N})$ is the space of probability distributions over $\mathbb{N}$. Also the effects of the operations are uncertain, $e : \mathcal{S} \times \mathcal{O} \to \Delta(\mathcal{S})$ and the initial states of the resources can be stochastic, as well, $i : \mathcal{R} \to \Delta(\mathcal{S})$. Note that the ranges of functions $d$, $e$ and $i$ contain probability distributions, we denote the corresponding random variables by $D$, $E$ and $I$, respectively. The notation $X \sim f$ indicate that random variable $X$ has probability distribution $f$. Thus, $D(s, o) \sim d(s, o)$, $E(s, o) \sim e(s, o)$ and $I(r) \sim i(r)$ for all $s \in \mathcal{S}$, $o \in \mathcal{O}$ and $r \in \mathcal{R}$. Take a look at Figure 4 for an illustration of the stochastic variants of the JSP and TSP problems.

### 2.3.1 Stochastic Dominance

In stochastic RAPs the performance of a solution is also a random variable. Therefore, in order to compare the performance of different solutions, we have to compare random variables. Many ways are known to make this comparison. We may say, for example, that a random variable has stochastic dominance over another random variable "almost surely", "in likelihood ratio sense", "stochastically", "in the increasing convex sense" or "in expectation". In different applications different types of comparisons may be suitable, however, probably the most natural one is based upon the expected values of the random variables. The paper applies this kind of comparison for stochastic RAPs.

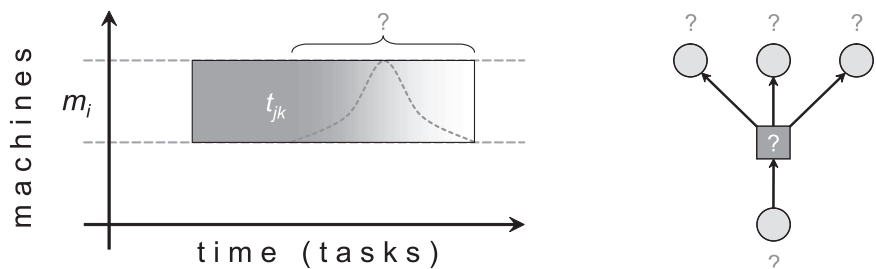

Figure 4: Randomization in case of JSP (left) and in case of TSP (right). In the latter, the initial state, the durations and the arrival vertex could be uncertain, as well.

### 2.3.2 Solution Classification

Now, we classify the basic types of resource allocation techniques. First, in order to give a proper classification we begin with recalling the concepts of "open-loop" and "closed-loop" controllers. An *open-loop* controller, also called a non-feedback controller, computes its input into a system by using only the current state and its model of the system. Therefore, an open-loop controller does not use feedback to determine if its input has achieved the desired goal, and it does not observe the output of the process being controlled. Conversely, a *closed-loop* controller uses feedback to control states or outputs of a dynamical system (Sontag, 1998). Closed-loop control has a significant advantage over open-loop solutions in dealing with *uncertainties*. Hence, it has improved reference tracking performance, it can stabilize unstable processes and reduced sensitivity to parameter variations.

In *deterministic* RAPs there is no significant difference between open- and closed-loop controls. In this case we can safely restrict ourselves to open-loop methods. If the solution is aimed at generating the resource allocation off-line in advance, then it is called *predictive*. Thus, predictive solutions perform open-loop control and assume a deterministic environment. In *stochastic* resource allocation there are some data (e.g., the actual durations) that will be available only during the execution of the plan. Based on the usage of this information, we identify two basic types of solution techniques. An open-loop solution that can deal with the uncertainties of the environment is called *proactive*. A proactive solution allocates the operations to resources and defines the orders of the operations, but, because the durations are uncertain, it does not determine precise starting times. This kind of technique can be applied only when the durations of the operations are stochastic, but, the states of the resources are known perfectly (e.g., stochastic JSP). Finally, in the stochastic case closed-loop solutions are called *reactive*. A reactive solution is allowed to make the decisions on-line, as the process actually evolves providing more information. Naturally, a reactive solution is not a simple sequence, but rather a resource allocation *policy* (to be defined later) which controls the process. The paper focuses on reactive solutions, only. We will formulate the reactive solution of a stochastic RAP as a control policy of a suitably defined Markov decision process (specially, a stochastic shortest path problem).

## 2.4 Markov Decision Processes

Sequential decision-making under the presence of uncertainties is often modeled by MDPs (Bertsekas & Tsitsiklis, 1996; Sutton & Barto, 1998; Feinberg & Shwartz, 2002). This section contains the basic definitions, the notations applied and some preliminaries.

By a (finite, discrete-time, stationary, fully observable) *Markov decision process* (MDP) we mean a stochastic system characterized by a 6-tuple $\langle \mathbb{X}, \mathbb{A}, \mathcal{A}, p, g, \alpha \rangle$, where the components are as follows: $\mathbb{X}$ is a finite set of discrete *states* and $\mathbb{A}$ is a finite set of control *actions*. Mapping $\mathcal{A} : \mathbb{X} \to \mathcal{P}(\mathbb{A})$ is the *availability function* that renders each state a set of actions available in the state where $\mathcal{P}$ denotes the power set. The *transition-probability* function is given by $p : \mathbb{X} \times \mathbb{A} \to \Delta(\mathbb{X})$, where $\Delta(\mathbb{X})$ is the space of probability distributions over $\mathbb{X}$. Let $p(y \,|\, x, a)$ denote the probability of arrival at state $y$ after executing action $a \in \mathcal{A}(x)$ in state $x$. The *immediate-cost* function is defined by $g : \mathbb{X} \times \mathbb{A} \to \mathbb{R}$, where $g(x, a)$ is the cost of taking action $a$ in state $x$. Finally, constant $\alpha \in [0, 1]$ denotes the *discount rate* or discount factor. If $\alpha = 1$, then the MDP is called *undiscounted*, otherwise it is *discounted*.

It is possible to extend the theory to more general state and action spaces, but at the expense of increased mathematical complexity. Finite state and action sets are mostly sufficient for digitally implemented controls and, therefore, we restrict ourselves to this case.

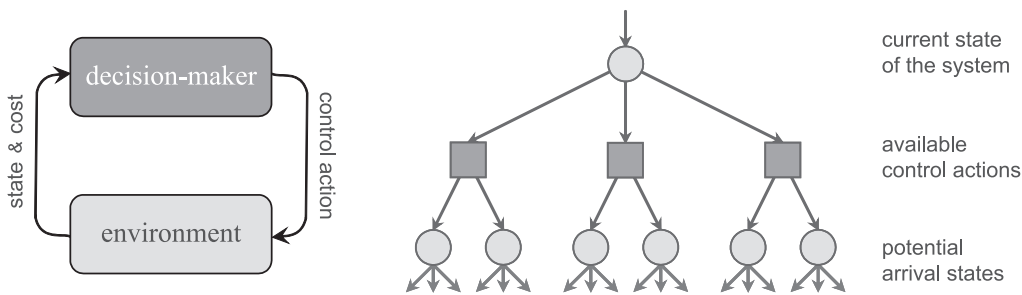

Figure 5: Markov decision processes – the interaction of the decision-maker and the uncertain environment (left); the temporal progress of the system (right).

An interpretation of an MDP can be given if we consider an *agent* that acts in an uncertain *environment*, which viewpoint is often taken in RL. The agent receives information about the state of the environment, $x$, in each state $x$ the agent is allowed to choose an action $a \in \mathcal{A}(x)$. After an action is selected, the environment moves to the next state according to the probability distribution $p(x, a)$, and the decision-maker collects its one-step cost, $g(x,a)$, as illustrated by Figure 5. The *aim* of the agent is to find an optimal behavior (policy) such that applying this strategy minimizes the expected cumulative costs.

A *stochastic shortest path* (SSP) problem is a special MDP in which the aim is to find a control policy such that reaches a pre-defined *terminal state* starting from a given initial state, additionally, minimizes the expected total costs of the path, as well. A policy is called *proper* if it reaches the terminal state with probability one. A usual assumption when dealing with SSP problems is that all policies are proper, abbreviated as APP.

### 2.4.1 Control Policies

A (stationary, Markov) control *policy* determines the action to take in each possible state. A *deterministic* policy, $\pi : \mathbb{X} \to \mathbb{A}$, is simply a function from states to control actions. A *randomized* policy, $\pi : \mathbb{X} \to \Delta(\mathbb{A})$, is a function from states to probability distributions over actions. We denote the probability of executing action $a$ in state $x$ by $\pi(x)(a)$ or, for short, by $\pi(x,a)$. Naturally, deterministic policies are special cases of randomized ones and, therefore, unless indicated otherwise, we consider randomized control policies.

For any $\widetilde{x}_0 \in \Delta(\mathbb{X})$ initial probability distribution of the states, the transition probabilities $p$ together with a control policy $\pi$ completely determine the progress of the system in a stochastic sense, namely, they define a *homogeneous Markov chain* on $\mathbb{X}$,

$$\widetilde{x}_{t+1} = P(\pi)\widetilde{x}_t,$$

where $\widetilde{x}_t$ is the state probability distribution vector of the system at time $t$, and $P(\pi)$ denotes the probability transition matrix induced by control policy $\pi$, formally defined as

$$[P(\pi)]_{x,y} = \sum_{a \in \mathbb{A}} p(y \,|\, x, a)\, \pi(x, a).$$

### 2.4.2 Value Functions

The *value* or *cost-to-go* function of a policy $\pi$ is a function from states to costs. It is defined on each state: $J^\pi : \mathbb{X} \to \mathbb{R}$. Function $J^\pi(x)$ gives the expected value of the cumulative (discounted) costs when the system is in state $x$ and it follows policy $\pi$ thereafter,

$$J^\pi(x) = \mathbb{E}\left[\sum_{t=0}^{N} \alpha^t g(X_t, A_t^\pi) \,\bigg|\, X_0 = x\right], \tag{1}$$

where $X_t$ and $A_t^\pi$ are random variables, $A_t^\pi$ is selected according to control policy $\pi$ and the distribution of $X_{t+1}$ is $p(X_t, A_t^\pi)$. The horizon of the problem is denoted by $N \in \mathbb{N} \cup \{\infty\}$. Unless indicated otherwise, we will always assume that the horizon is infinite, $N = \infty$.

Similarly to the definition of $J^\pi$, one can define *action-value* functions of control polices,

$$Q^\pi(x,a) = \mathbb{E}\left[\sum_{t=0}^{N} \alpha^t g(X_t, A_t^\pi) \,\bigg|\, X_0 = x, A_0^\pi = a\right],$$

where the notations are the same as in equation (1). Action-value functions are especially important for model-free approaches, such as the classical Q-learning algorithm.

### 2.4.3 BELLMAN EQUATIONS

We say that $\pi_1 \leq \pi_2$ if and only if, for all $x \in \mathbb{X}$, we have $J^{\pi_1}(x) \leq J^{\pi_2}(x)$. A control policy is (uniformly) optimal if it is less than or equal to all other control policies.

There always exists at least one optimal policy (Sutton & Barto, 1998). Although there may be many optimal policies, they all share the same unique optimal cost-to-go function, denoted by $J^*$. This function must satisfy the *Bellman optimality equation* (Bertsekas & Tsitsiklis, 1996), $TJ^* = J^*$, where $T$ is the *Bellman operator*, defined for all $x \in \mathbb{X}$, as

$$(TJ)(x) = \min_{a \in \mathcal{A}(x)} \Big[ g(x,a) + \alpha \sum_{y \in \mathbb{X}} p(y \,|\, x,a) J(y) \Big]. \tag{2}$$

The *Bellman equation* for an arbitrary (stationary, Markov, randomized) policy is

$$(T^{\pi}J)(x) = \sum_{a \in \mathcal{A}(x)} \pi(x,a) \Big[ g(x,a) + \alpha \sum_{y \in \mathbb{X}} p(y \,|\, x,a) J(y) \Big],$$

where the notations are the same as in equation (2) and we also have $T^{\pi} J^{\pi} = J^{\pi}$.

From a given value function $J$, it is straightforward to get a policy, e.g., by applying a *greedy* and deterministic policy (w.r.t. $J$) that always selects actions of minimal costs,

$$\pi(x) \in \arg\min_{a \in \mathcal{A}(x)} \Big[ g(x,a) + \alpha \sum_{y \in \mathbb{X}} p(y \,|\, x,a) J(y) \Big].$$

MDPs have an extensively studied theory and there exist a lot of exact and approximate solution methods, e.g., value iteration, policy iteration, the Gauss-Seidel method, Q-learning, Q($\lambda$), SARSA and TD($\lambda$) - temporal difference learning (Bertsekas & Tsitsiklis, 1996; Sutton & Barto, 1998; Feinberg & Shwartz, 2002). Most of these reinforcement learning algorithms work by iteratively approximating the optimal value function.

## 2.5 Reactive Resource Control

In this section we formulate reactive solutions of stochastic RAPs as control policies of suitably reformulated SSP problems. The current task durations and resource states will only be incrementally available during the resource allocation control process.

### 2.5.1 PROBLEM REFORMULATION

A state $x \in \mathbb{X}$ is defined as a 4-tuple $x = \langle \tau, \mu, \varrho, \varphi \rangle$, where $\tau \in \mathbb{N}$ is the current time and the function $\mu : \mathcal{R} \to \mathcal{S}$ determines the current states of the resources. The *partial* functions $\varrho$ and $\varphi$ store the past of the process, namely, $\varrho : \mathcal{R} \times \mathbb{N}_{\tau-1} \hookrightarrow \mathcal{O}$ contains the resources and the times in which an operation was started and $\varphi : \mathcal{R} \times \mathbb{N}_{\tau-1} \hookrightarrow \mathbb{N}_{\tau}$ describes the finish times of the already completed operations, where $\mathbb{N}_{\tau} = \{0, \dots, \tau\}$. Naturally, $dom(\varphi) \subseteq dom(\varrho)$. By $\mathcal{T}_S(x) \subseteq \mathcal{T}$ we denote the set of tasks which have been started in state $x$ (before the current time $\tau$) and by $\mathcal{T}_F(x) \subseteq \mathcal{T}_S(x)$ the set of tasks that have been finished already in state $x$. It is easy to see that $\mathcal{T}_S(x) = rng(\varrho) \cap \mathcal{T}$ and $\mathcal{T}_F(x) = rng(\varrho|_{dom(\varphi)}) \cap \mathcal{T}$, where $rng(\cdot)$ denotes the range or image set of a function. The process starts from an initial state

$x_s = \langle 0, \mu, \emptyset, \emptyset \rangle$, which corresponds to the situation at time zero when none of the operations have been started. The initial probability distribution, $\beta$, can be calculated as follows

$$\beta(x_s) = \mathbb{P}\left(\mu(r_1) = I(r_1), \ldots, \mu(r_n) = I(r_n)\right),$$

where $I(r) \sim i(r)$ denotes the random variable that determines the initial state of resource $r \in \mathcal{R}$ and $n = |\mathcal{R}|$. Thus, $\beta$ renders initial states to resources according to the (multivariate) probability distribution $I$ that is a component of the RAP. We introduce a set of terminal states, as well. A state $x$ is considered as a terminal state ($x \in \mathbb{T}$) in two cases. First, if all the tasks are finished in the state, formally, if $\mathcal{T}_F(x) = \mathcal{T}$ and it can be reached from a state $\hat{x}$, where $\mathcal{T}_F(\hat{x}) \neq \mathcal{T}$. Second, if the system reached a state where no tasks or operations can be executed, in other words, if the allowed set of actions is empty, $\mathcal{A}(x) = \emptyset$.

It is easy to see that, in theory, we can *aggregate* all terminal states to a global unique terminal state and introduce a new unique initial state, $x_0$, that has only one available action which takes us randomly (with $\beta$ distribution) to the real initial states. Then, the problem becomes a *stochastic shortest path* problem and the aim can be described as finding a routing having minimal expected cost from the new initial state to the goal state.

At every time $\tau$ the system is informed on the finished operations, and it can decide on the operations to apply (and by which resources). The control action space contains operation-resource assignments $a_{vr} \in \mathbb{A}$, where $v \in \mathcal{O}$ and $r \in \mathcal{R}$, and a special $a_{wait}$ control that corresponds to the action when the system does not start a new operation at the current time. In a non-terminal state $x = \langle \tau, \mu, \varrho, \varphi \rangle$ the available actions are

$$a_{wait} \in \mathcal{A}(x) \Leftrightarrow \mathcal{T}_S(x) \setminus \mathcal{T}_F(x) \neq \emptyset$$

$$\forall v \in \mathcal{O} : \forall r \in \mathcal{R} : a_{vr} \in \mathcal{A}(x) \Leftrightarrow (v \in \mathcal{O} \setminus \mathcal{T}_S(x) \ \wedge \ \forall \langle \hat{r}, t \rangle \in dom(\varrho) \setminus dom(\varphi) : \hat{r} \neq r \ \wedge$$

$$\wedge \ \langle \mu(r), v \rangle \in dom(d) \ \wedge \ v \in \mathcal{T} \Rightarrow (\forall u \in \mathcal{T} : \langle u, v \rangle \in \mathcal{C} \Rightarrow u \in \mathcal{T}_F(x)))$$

Thus, action $a_{wait}$ is available in every state with an unfinished operation; action $a_{vr}$ is available in states in which resource $r$ is idle, it can process operation $v$, additionally, if $v$ is a task, then it was not executed earlier and its precedence constraints are satisfied.

If an action $a_{vr} \in \mathcal{A}(x)$ is executed in a state $x = \langle \tau, \mu, \varrho, \varphi \rangle$, then the system moves with probability one to a new state $\hat{x} = \langle \tau, \mu, \hat{\varrho}, \varphi \rangle$, where $\hat{\varrho} = \varrho \cup \{\langle \langle r, t \rangle, v \rangle\}$. Note that we treat functions as sets of ordered pairs. The resulting $\hat{x}$ corresponds to the state where operation $v$ has started on resource $r$ if the previous state of the environment was $x$.

The effect of the $a_{wait}$ action is that from $x = \langle \tau, \mu, \varrho, \varphi \rangle$ it takes to an $\hat{x} = \langle \tau+1, \hat{\mu}, \varrho, \hat{\varphi} \rangle$, where an unfinished operation $\varrho(r, t)$ that was started at $t$ on $r$ finishes with probability

$$\mathbb{P}(\langle r, t \rangle \in dom(\hat{\varphi}) \mid x, \langle r, t \rangle \in dom(\varrho) \setminus dom(\varphi)) = \frac{\mathbb{P}(D(\mu(r), \varrho(r,t)) + t = \tau)}{\mathbb{P}(D(\mu(r), \varrho(r,t)) + t \geq \tau)},$$

where $D(s, v) \sim d(s, v)$ is a random variable that determines the duration of operation $v$ when it is executed by a resource which has state $s$. This quantity is called *completion rate* in stochastic scheduling theory and *hazard rate* in reliability theory. We remark that for operations with continuous durations, this quantity is defined by $f(t)/(1 - F(t))$, where $f$ denotes the density function and $F$ the distribution of the random variable that determines the duration of the operation. If operation $v = \varrho(r, t)$ has finished ($\langle r, t \rangle \in dom(\hat{\varphi})$), then $\hat{\varphi}(r, t) = \tau$ and $\hat{\mu}(r) = E(r, v)$, where $E(r, v) \sim e(r, v)$ is a random variable that determines the new state of resource $r$ after it has executed operation $v$. Except the extension of its domain set, the other values of function $\varphi$ do not change, consequently, $\forall \langle r, t \rangle \in dom(\varphi)$ : $\hat{\varphi}(r, t) = \varphi(r, t)$. In other words, $\hat{\varphi}$ is a conservative extension of $\varphi$, formally, $\varphi \subseteq \hat{\varphi}$.

The immediate-cost function $g$ for a given $\kappa$ performance measure is defined as follows. Assume that $\kappa$ depends only on the operation-resource assignments and the completion times. Let $x = \langle \tau, \mu, \varrho, \varphi \rangle$ and $\hat{x} = \langle \hat{\tau}, \hat{\mu}, \hat{\varrho}, \hat{\varphi} \rangle$. Then, in general, if the system arrives at state $\hat{x}$ after executing control action $a$ in state $x$, it incurs cost $\kappa(\varrho, \varphi) - \kappa(\hat{\varrho}, \hat{\varphi})$.

Note that, though, in Section 2.2.2 performance measures were defined on complete solutions, for most measures applied in practice, such as total completion time or weighted total lateness, it is straightforward to generalize the performance measure to partial solutions, as well. One may, for example, treat the partial solution of a problem as a complete solution of a smaller (sub)problem, namely, a problem with fewer tasks to be completed.

If the control process has failed, more precisely, if it was not possible to finish all tasks, then the immediate-cost function should render penalties (depending on the specific problem) regarding the non-completed tasks proportional to the number of these failed tasks.

### 2.5.2 FAVORABLE FEATURES

Let us call the introduced SSPs, which describe stochastic RAPs, *RAP-MDPs*. In this section we overview some basic properties of RAP-MDPs. First, it is straightforward to see that these MDPs have *finite* action spaces, since $|\mathbb{A}| \leq |\mathcal{R}| \, |\mathcal{O}| + 1$ always holds.

Though, the state space of a RAP-MDP is denumerable in general, if the allowed number of non-task operations is bounded and the random variables describing the operation durations are finite, the state space of the reformulated MDP becomes *finite*, as well.

We may also observe that RAP-MDPs are *acyclic* (or aperiodic), viz., none of the states can appear multiple times, because during the resource allocation process $\tau$ and $dom(\varrho)$ are non-decreasing and, additionally, each time the state changes, the quantity $\tau + |dom(\varrho)|$ strictly increases. Therefore, the system cannot reach the same state twice. As an immediate consequence, all policies eventually terminate (if the MDP was finite) and, thus, are proper.

For the effective computation of a good control policy, it is important to try to reduce the number of states. We can do so by recognizing that if the performance measure $\kappa$ is non-decreasing in the completion times, then an optimal control policy of the reformulated RAP-MDP can be found among the policies which start new operations only at times when another operation has been finished or in an initial state. This statement can be supported by the fact that without increasing the cost ($\kappa$ is non-decreasing) every operation can be shifted earlier on the resource which was assigned to it until it reaches another operation, or until it reaches a time when one of its preceding tasks is finished (if the operation was a task with precedence constrains), or, ultimately, until time zero. Note that most of the performance measures used in practice (e.g., makespan, weighted completion time, average tardiness) are non-decreasing. As a consequence, the states in which no operation has been finished can be omitted, except the initial states. Therefore, each $a_{wait}$ action may lead to a state where an operation has been finished. We may consider it, as the system executes automatically an $a_{wait}$ action in the omitted states. By this way, the state space can be decreased and, therefore, a good control policy can be calculated more effectively.

### 2.5.3 Composable Measures

For a large class of performance measures, the state representation can be simplified by leaving out the past of the process. In order to do so, we must require that the performance measure be composable with a suitable function. In general, we call a function $f : \mathcal{P}(X) \to \mathbb{R}$ $\gamma$-*composable* if for any $A, B \subseteq X$, $A \cap B = \emptyset$ it holds that $\gamma(f(A), f(B)) = f(A \cup B)$, where $\gamma : \mathbb{R} \times \mathbb{R} \to \mathbb{R}$ is called the *composition function*, and $X$ is an arbitrary set. This definition can be directly applied to performance measures. If a performance measure, for example, is $\gamma$-composable, it indicates that the value of any complete solution can be computed from the values of its disjoint subsolutions (solutions to subproblems) with function $\gamma$. In practical situations the composition function is often the max, the min or the "+" function.

If the performance measure $\kappa$ is $\gamma$-composable, then the past can be omitted from the state representation, because the performance can be calculated incrementally. Thus, a state can be described as $x = \langle \bar{\tau}, \bar{\kappa}, \bar{\mu}, \mathcal{T}_U \rangle$, where $\bar{\tau} \in \mathbb{N}$, as previously, is the current time, $\bar{\kappa} \in \mathbb{R}$ contains the performance of the current (partial) solution and $\mathcal{T}_U$ is the set of unfinished tasks. The function $\bar{\mu} : \mathcal{R} \to \mathcal{S} \times (\mathcal{O} \cup \{\iota\}) \times \mathbb{N}$ determines the current states of the resources together with the operations currently executed by them (or $\iota$ if a resource is idle) and the starting times of the operations (needed to compute their completion rates).

In order to keep the analysis as simple as possible, we restrict ourselves to composable functions, since almost all performance measures that appear in practice are $\gamma$-composable for a suitable $\gamma$ (e.g., makespan or total production time is max-composable).

### 2.5.4 Reactive Solutions

Now, we are in a position to define the concept of reactive solutions for stochastic RAPs. A *reactive solution* is a (stationary, Markov) control policy of the reformulated SSP problem. A reactive solution performs a closed-loop control, since at each time step the controller is informed about the current state of system and it can choose a control action based upon this information. Section 3 deals with the computation of effective control policies.

## 3. Solution Methods

In this section we aim at giving an effective solution to large-scale RAPs in uncertain and changing environments with the help of different machine learning approaches. First, we overview some approximate dynamic programming methods to compute a "good" policy. Afterwards, we investigate two function approximation techniques to enhance the solution. Clustering, rollout algorithm and action space decomposition as well as distributed sampling are also considered, as they can speedup the computation of a good control policy considerably and, therefore, are important additions if we face large-scale problems.

### 3.1 Approximate Dynamic Programming

In the previous sections we have formulated RAPs as acyclic (aperiodic) SSP problems. Now, we face the challenge of finding a good policy. In theory, the optimal value function of a finite MDP can be exactly computed by *dynamic programming* (DP) methods, such as value iteration or the Gauss-Seidel method. Alternatively, an exact optimal policy can be directly calculated by policy iteration. However, due to the "curse of dimensionality", computing an exact optimal solution by these methods is practically infeasible, e.g., typically both the required amount of computation and the needed storage space, viz., memory, grows quickly with the size of the problem. In order to handle the "curse", we should apply *approximate dynamic programming* (ADP) techniques to achieve a good approximation of an optimal policy. Here, we suggest using sampling-based *fitted Q-learning* (FQL). In each trial a Monte-Carlo estimate of the value function is computed and projected onto a suitable function space. The methods described in this section (FQL, MCMC and the Boltzmann formula) should be applied simultaneously, in order to achieve an efficient solution.

### 3.1.1 Fitted Q-learning

Watkins' Q-learning is a very popular off-policy model-free reinforcement learning algorithm (Even-Dar & Mansour, 2003). It works with action-value functions and iteratively approximates the optimal value function. The one-step Q-learning rule is defined as follows

$$Q_{i+1}(x, a) = (1 - \gamma_i(x, a))Q_i(x, a) + \gamma_i(x, a)(\widetilde{T}_i Q_i)(x, a),$$

$$(\widetilde{T}_i Q_i)(x, a) = g(x, a) + \alpha \min_{B \in \mathcal{A}(Y)} Q_i(Y, B),$$

where $\gamma_i(x, a)$ are the learning rates and $Y$ is a random variable representing a state generated from the pair $(x, a)$ by simulation, that is, according to the probability distribution $p(x, a)$. It is known (Bertsekas & Tsitsiklis, 1996) that if $\gamma_i(x) \in [0, 1]$ and they satisfy

$$\sum_{i=0}^{\infty} \gamma_i(x, a) = \infty \quad \text{and} \quad \sum_{i=0}^{\infty} \gamma_i^2(x, a) < \infty,$$

then the Q-learning algorithm converges with probability one to the optimal action-value function, $Q^*$, in the case of lookup table representation when each state-action value is stored independently. We speak about the method of *fitted Q-learning* (FQL) when the value function is represented by a (typically parametric) function from a suitable function space, $\mathcal{F}$, and after each iteration, the updated value function is projected back onto $\mathcal{F}$.

A useful observation is that we need the "learning rate" parameters only to overcome the effects of random disturbances. However, if we deal with deterministic problems, this part of the method can be simplified. The resulting algorithm simply updates $Q(x, a)$ with the minimum of the previously stored estimation and the current outcome of the simulation, which is also the core idea of the LRTA* algorithm (Bulitko & Lee, 2006). When we dealt with deterministic resource allocation problems, we applied this simplification, as well.

### 3.1.2 Evaluation by Simulation

Naturally, in large-scale problems we cannot update all states at once. Therefore, we perform *Markov chain Monte Carlo* (MCMC) simulations (Hastings, 1970; Andrieu, Freitas, Doucet, & Jordan, 2003) to generate samples with the model, which are used for computing the new approximation of the estimated cost-to-go function. Thus, the set of states to be updated in episode $i$, namely $\mathbb{X}_i$, is generated by simulation. Because RAP-MDPs are acyclic, we apply *prioritized sweeping*, which means that after each iteration, the cost-to-go estimations are updated in the reverse order in which they appeared during the simulation.

Assume, for example, that $\mathbb{X}_i = \left\{ x_1^i, x_2^i, \ldots, x_{t_i}^i \right\}$ is the set of states for the update of the value function after iteration $i$, where $j < k$ implies that $x_j^i$ appeared earlier during the simulation than $x_k^i$. In this case the order in which the updates are performed, is $x_{t_i}^i, \ldots, x_1^i$. Moreover, we do not need a *uniformly* optimal value function, it is enough to have a good approximation of the optimal cost-to-go function for the relevant states. A state is called *relevant* if it can appear with positive probability during the application of an optimal policy. Therefore, it is sufficient to consider the case when $x_1^i = x_0$, where $x_1^i$ is the first state in episode $i$ and $x_0$ is the (aggregated) initial state of the SSP problem.

### 3.1.3 The Boltzmann Formula

In order to ensure the convergence of the FQL algorithm, one must guarantee that each cost-to-go estimation be continuously updated. A technique used often to balance between *exploration* and *exploitation* is the *Boltzmann formula* (also called *softmin* action selection):

$$\pi_i(x, a) = \frac{\exp(-Q_i(x, a)/\tau)}{\sum\limits_{b \in \mathcal{A}(x)} \exp(-Q_i(x, b)/\tau)},$$

where $\tau \geq 0$ is the Boltzmann (or Gibbs) temperature, $i$ is the episode number. It is easy to see that high temperatures cause the actions to be (nearly) equiprobable, low ones cause a greater difference in selection probability for actions that differ in their value estimations. Note that here we applied the Boltzmann formula for minimization, viz., small values result in high probability. It is advised to extend this approach by a variant of *simulated annealing* (Kirkpatrick, Gelatt, & Vecchi, 1983) or *Metropolis algorithm* (Metropolis, Rosenbluth, Rosenbluth, Teller, & Teller, 1953), which means that $\tau$ should be decreased over time, at a suitable, e.g., logarithmic, rate (Singh, Jaakkola, Littman, & Szepesvári, 2000).

## 3.2 Cost-to-Go Representations

In Section 3.1 we suggested FQL for iteratively approximating the optimal value function. However, the question of a suitable function space, onto which the resulted value functions can be effectively projected, remained open. In order to deal with large-scale problems (or problems with continuous state spaces) this question is crucial. In this section, first, we suggest features for stochastic RAPs, then describe two methods that can be applied to compactly represent value functions. The first and simpler one applies hash tables while the second, more sophisticated one, builds upon the theory of support vector machines.

### 3.2.1 Feature Vectors

In order to efficiently apply a function approximator, first, the states and the actions of the reformulated MDP should be associated with numerical vectors representing, e.g., typical features of the system. In the case of stochastic RAPs, we suggest using features as follows:

- For each resource in $\mathcal{R}$, the *resource state id*, the *operation id* of the operation being currently processed by the resource (could be idle), as well as the *starting time* of the last (and currently unfinished) operation can be a feature. If the model is available to the system, the *expected ready time* of the resource should be stored instead.

- For each task in $\mathcal{T}$, the *task state id* could be treated as a feature that can assume one of the following values: "not available" (e.g., some precedence constraints are not satisfied), "ready for execution", "being processed" or "finished". It is also advised to apply "1-out-of-n" coding, viz., each value should be associated with a separate bit.

- In case we use action-value functions, for each action (resource-operation assignment) the *resource id* and the *operation id* could be stored. If the model is available, then the *expected finish time* of the operation should also be taken into account.

In the case of a model-free approach which applies action-value functions, for example, the feature vector would have $3 \cdot |\mathcal{R}| + |\mathcal{T}| + 2$ components. Note that for features representing temporal values, it is advised to use *relative* time values instead of *absolute* ones.

### 3.2.2 HASH TABLES

Suppose that we have a vector $w = \langle w_1, w_2, \ldots, w_k \rangle$, where each component $w_i$ corresponds to a feature of a state or an action. Usually, the value estimations for all of these vectors cannot be stored in the memory. In this case one of the simplest methods to be applied is to represent the estimations in a *hash table*. A hash table is, basically, a dictionary in which keys are mapped to array positions by hash functions. If all components can assume finite values, e.g., in our finite-state, discrete-time case, then a key could be generated as follows. Let us suppose that for all $w_i$ we have $0 \le w_i < m_i$, then $w$ can be seen as a number in a *mixed radix numeral system* and, therefore, a unique *key* can be calculated as

$$\varphi(w) = \sum_{i=1}^{k} w_i \prod_{j=1}^{i-1} m_j,$$

where $\varphi(w)$ denotes the key of $w$, and the value of an empty product is treated as one.

The *hash function*, $\psi$, maps feature vector keys to memory positions. More precisely, if we have memory for storing only $d$ value estimations, then the hash function takes the form $\psi : rng(\varphi) \to \{0, \ldots, d-1\}$, where $rng(\cdot)$ denotes the range set of a function.

It is advised to apply a $d$ that is *prime*. In this case a usual hashing function choice is $\psi(x) = y$ if and only if $y \equiv x \pmod{d}$, namely, if $y$ is congruent to $x$ modulo $d$.

Having the keys of more than one item map to the same position is called a *collision*. In the case of RAP-MDPs we suggest a collision resolution method as follows. Suppose that during a value update the feature vector of a state (or a state-action pair) maps to a position that is already occupied by another estimation corresponding to another item (which can be detected, e.g., by storing the keys). Then we have a collision and the estimation of the new item should overwrite the old estimation if and only if the MDP state corresponding to the new item appears with higher probability during execution starting from the (aggregated) initial state than the one corresponding to the old item. In case of a model-free approach, the item having a state with smaller current time component can be kept.

Despite its simplicity, the hash table representation has several disadvantages, e.g., it still needs a lot of memory to work efficiently, it cannot easily handle continuous values and, it only stores individual data, moreover, it does not *generalize* to "similar" items. In the next section we present a statistical approach that can deal with these issues, as well.

### 3.2.3 Support Vector Regression

A promising choice for compactly representing the cost-to-go function is to use *support vector regression* (SVR) from statistical learning theory. For maintaining the value function estimations, we suggest using $\nu$-SVRs which were proposed by Schölkopf, Smola, Williamson, and Bartlett (2000). They have an advantage over classical $\varepsilon$-SVRs according to which, through the new parameter $\nu$, the number of support vectors can be controlled. Additionally, parameter $\varepsilon$ can be eliminated. First, the core ideas of $\nu$-SVRs are presented.

In general, SVR addresses the problem as follows. We are given a sample, a set of data points $\{\langle x_1, y_1 \rangle, \ldots, \langle x_l, y_l \rangle\}$, such that $x_i \in \mathcal{X}$ is an input, where $\mathcal{X}$ is a measurable space, and $y_i \in \mathbb{R}$ is the target output. For simplicity, we shall assume that $\mathcal{X} \subseteq \mathbb{R}^k$, where $k \in \mathbb{N}$. The aim of the learning process is to find a function $f : \mathcal{X} \to \mathbb{R}$ with a small risk

$$R[f] = \int_{\mathcal{X}} l(f, x, y) dP(x, y),  \tag{3}$$

where $P$ is a probability measure, which is responsible for the generation of the observations and $l$ is a loss function, such as $l(f, x, y) = (f(x) - y)^2$. A common error function used in SVRs is the so-called $\varepsilon$-*insensitive* loss function, $|f(x) - y|_\varepsilon = \max\{0, |f(x) - y| - \varepsilon\}$. Unfortunately, we cannot minimize (3) directly, since we do not know $P$, we are given the sample, only (generated, e.g., by simulation). We try to obtain a small risk by minimizing the regularized risk functional in which we average over the training sample

$$\frac{1}{2} \|w\|^2 + C \cdot R_{emp}^\varepsilon[f],  \tag{4}$$

where, $\|w\|^2$ is a term that characterizes the model complexity and $C > 0$ a constant that determines the trade-off between the flatness of the regression and the amount up to which deviations larger than $\varepsilon$ are tolerated. The function $R_{emp}^\varepsilon[f]$ is defined as follows

$$R_{emp}^\varepsilon[f] = \frac{1}{l} \sum_{i=1}^{l} |f(x_i) - y_i|_\varepsilon.$$

It measures the $\varepsilon$-insensitive average training error. The problem which arises when we try to minimize (4) is called *empirical risk minimization* (ERM). In regression problems we usually have a Hilbert space $\mathcal{F}$, containing $\mathcal{X} \to \mathbb{R}$ type (typically non-linear) functions, and our aim is to find a function $f$ that is "close" to $y_i$ in each $x_i$ and takes the form

$$f(x) = \sum_j w_j \phi_j(x) + b = w^T \phi(x) + b,$$

where $\phi_j \in \mathcal{F}$, $w_j \in \mathbb{R}$ and $b \in \mathbb{R}$. Using Lagrange multiplier techniques, we can rewrite the regression problem in its dual form (Schölkopf et al., 2000) and arrive at the final $\nu$-SVR optimization problem. The resulting regression estimate then takes the form as follows

$$f(x) = \sum_{i=1}^{l} (\alpha_i^* - \alpha_i) K(x_i, x) + b,$$

where $\alpha_i$ and $\alpha_i^*$ are the Lagrange multipliers, and $K$ denotes an *inner product kernel* defined by $K(x, y) = \langle \phi(x), \phi(y) \rangle$, where $\langle \cdot, \cdot \rangle$ denotes inner product. Note that $\alpha_i, \alpha_i^* \neq 0$ holds usually only for a small subset of training samples, furthermore, parameter $b$ (and $\varepsilon$) can be determined as well, by applying the Karush-Kuhn-Tucker (KKT) conditions.

Mercer's theorem in functional analysis characterizes which functions correspond to an inner product in some space $\mathcal{F}$. Basic kernel types include linear, polynomial, Gaussian and sigmoid functions. In our experiments with RAP-MDPs we have used Gaussian kernels which are also called *radial basis function* (RBF) kernels. RBF kernels are defined by $K(x, y) = \exp(-\|x - y\|^2 / (2\sigma^2))$, where $\sigma > 0$ is an adjustable kernel parameter.

A variant of the fitted Q-learning algorithm combined with regression and softmin action selection is described in Table 1. It simulates a state-action trajectory with the model and updates the estimated values of only the state-action pairs which appeared during the simulation. Most of our RAP solutions described in the paper are based on this algorithm.

The notations of the pseudocode shown in Table 1 are as follows. Variable $i$ contains the episode number, $t_i$ is the length of episode $i$ and $j$ is a parameter for time-steps inside an episode. The Boltzmann temperature is denoted by $\tau$, $\pi_i$ is the control policy applied in episode $i$ and $x_0$ is the (aggregated) initial state. State $x_j^i$ and action $a_j^i$ correspond to step $j$ in episode $i$. Function $h$ computes features for state-action pairs while $\gamma_i$ denotes learning rates. Finally, $\mathcal{L}_i$ denotes the regression sample and $Q_i$ is the fitted value function.

Although, support vector regression offers an elegant and efficient solution to the value function representation problem, we presented the hash table representation possibility not only because it is much easier to implement, but also because it requires less computation, thus, provides faster solutions. Moreover, the values of the hash table could be accessed independently; this was one of the reasons why we applied hash tables when we dealt with distributed solutions, e.g., on architectures with uniform memory access. Nevertheless, SVRs have other advantages, most importantly, they can "generalize" to "similar" data.

| **Regression Based Q-learning** |
|---|
| 1.  **Initialize** $Q_0$, $\mathcal{L}_0$, $\tau$ and let $i = 1$. |
| 2.  **Repeat** (for each episode) |
| 3.      **Set** $\pi_i$ to a soft and semi-greedy policy w.r.t. $Q_{i-1}$, e.g., <br>       $\pi_i(x, a) = \exp(-Q_{i-1}(x, a)/\tau) / \left[\sum_{b \in \mathcal{A}(x)} \exp(-Q_{i-1}(x, b)/\tau)\right].$ |
| 4.      **Simulate** a state-action trajectory from $x_0$ using policy $\pi_i$. |
| 5.      **For** $j = t_i$ **to** 1 (for each state-action pair in the episode) **do** |
| 6.          **Determine** the features of the state-action pair, $y_j^i = h(x_j^i, a_j^i)$. |
| 7.          **Compute** the new action-value estimation for $x_j^i$ and $a_j^i$, e.g., <br>           $z_j^i = (1 - \gamma_i)Q_{i-1}(x_j^i, a_j^i) + \gamma_i \left[g(x_j^i, a_j^i) + \alpha \min_{b \in \mathcal{A}(x_{j+1}^i)} Q_{i-1}(x_{j+1}^i, b)\right].$ |
| 8.      **End loop** (end of state-action processing) |
| 9.      **Update** sample set $\mathcal{L}_{i-1}$ with $\left\{\langle y_j^i, z_j^i \rangle : j = 1, \ldots, t_i\right\}$, the result is $\mathcal{L}_i$. |
| 10.     **Calculate** $Q_i$ by fitting a smooth regression function to the sample of $\mathcal{L}_i$. |
| 11.     **Increase** the episode number, $i$, and **decrease** the temperature, $\tau$. |
| 12.  **Until** some terminating conditions are met, e.g., $i$ reaches a limit <br>      or the estimated approximation error to $Q^*$ gets sufficiently small. |
| **Output:** the action-value function $Q_i$ (or $\pi(Q_i)$, e.g., the greedy policy w.r.t. $Q_i$). |

Table 1: Pseudocode for regression-based Q-learning with softmin action selection.

### 3.3 Additional Improvements

Computing a (close-to) optimal solution with RL methods, such as (fitted) Q-learning, could be very inefficient in large-scale systems, even if we apply prioritized sweeping and a capable representation. In this section we present some additional improvements in order to speed up to optimization process, even at the expense of achieving only suboptimal solutions.

#### 3.3.1 Rollout Algorithms

During our experiments, which are presented in Section 4, it turned out that using a *sub-optimal base policy*, such as a greedy policy with respect to the immediate costs, to guide the exploration, speeds up the optimization considerably. Therefore, at the initial stage we suggest applying a *rollout policy*, which is a limited lookahead policy, with the optimal cost-to-go approximated by the cost-to-go of the base policy (Bertsekas, 2001). In order to introduce the concept more precisely, let $\hat{\pi}$ be the greedy policy w.r.t. immediate-costs,

$$\hat{\pi}(x) \in \operatorname*{arg\,min}_{a \in \mathcal{A}(x)} g(x, a).$$

The value function of $\hat{\pi}$ is denoted by $J^{\hat{\pi}}$. The one-step lookahead rollout policy $\pi$ based on policy $\hat{\pi}$, which is an improvement of $\hat{\pi}$ (cf. policy iteration), can be calculated by

$$\pi(x) \in \operatorname*{arg\,min}_{a \in \mathcal{A}(x)} \mathbb{E}\left[G(x, a) + J^{\hat{\pi}}(Y)\right],$$

where $Y$ is a random variable representing a state generated from the pair $(x, a)$ by simulation, that is, according to probability distribution $p(x, a)$. The expected value (viz., the expected costs and the cost-to-go of the base policy) is approximated by Monte Carlo simulation of several trajectories that start at the current state. If the problem is deterministic, then a single simulation trajectory suffices, and the calculations are greatly simplified.

Take a look at Figure 6 for an illustration. In scheduling theory, a similar (but simplified) concept can be found and a rollout policy would be called a *dispatching rule*.

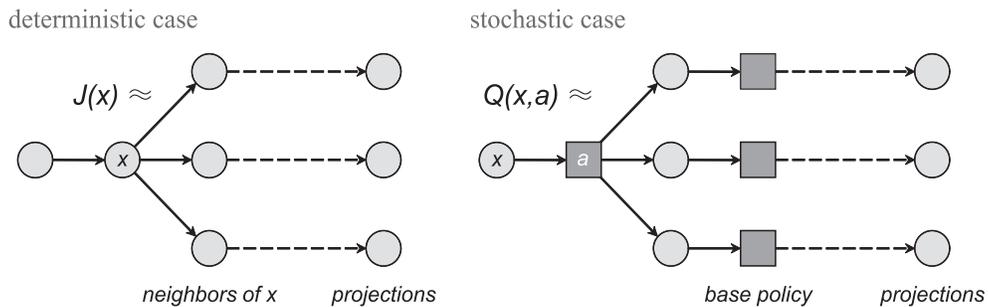

Figure 6: The evaluation of state $x$ with rollout algorithms in the deterministic (left) and the stochastic (right) case. Circles denote states and rectangles represent actions.

The two main issues why we suggest the application of rollout algorithms in the initial stages of value function approximation-based reinforcement learning are as follows:

1. We need several initial samples before the first application of approximation techniques and these first samples can be generated by simulations guided by a rollout policy.

2. General reinforcement learning methods perform quite poorly in practice without any initial guidance. However, the learning algorithm can start improving the rollout policy $\pi$, especially, in case we apply (fitted) Q-learning, it can learn directly from the trajectories generated by a rollout policy, since it is an *off-policy* learning method.

Our numerical experiments showed that rollout algorithms provide significant speedup.

### 3.3.2 ACTION SPACE DECOMPOSITION

In large-scale problems the set of available actions in a state may be very large, which can slow down the system significantly. In the current formulation of the RAP the number of available actions in a state is $O(|\mathcal{T}||\mathcal{R}|)$. Though, even in real world situations $|\mathcal{R}|$ is, usually, not very large, but $\mathcal{T}$ could contain thousands of tasks. Here, we suggest decomposing the action space as shown in Figure 7. First, the system selects a task, only, and it moves to a new state where this task is fixed and an executing resource should be selected. In this case the state description can be extended by a new variable $\tau \in \mathcal{T} \cup \{\emptyset\}$, where $\emptyset$ denotes the case when no task has been selected yet. In every other case the system should select an executing resource for the selected task. Consequently, the new action space is $\mathbb{A} = \mathbb{A}_1 \cup \mathbb{A}_2$, where $\mathbb{A}_1 = \{a_v \mid v \in \mathcal{T}\} \cup \{a_\omega\}$ and $\mathbb{A}_2 = \{a_r \mid r \in \mathcal{R}\}$. As a result, we radically decreased the number of available actions, however, the number of possible states was increased. Our experiments showed that it was a reasonable trade-off.

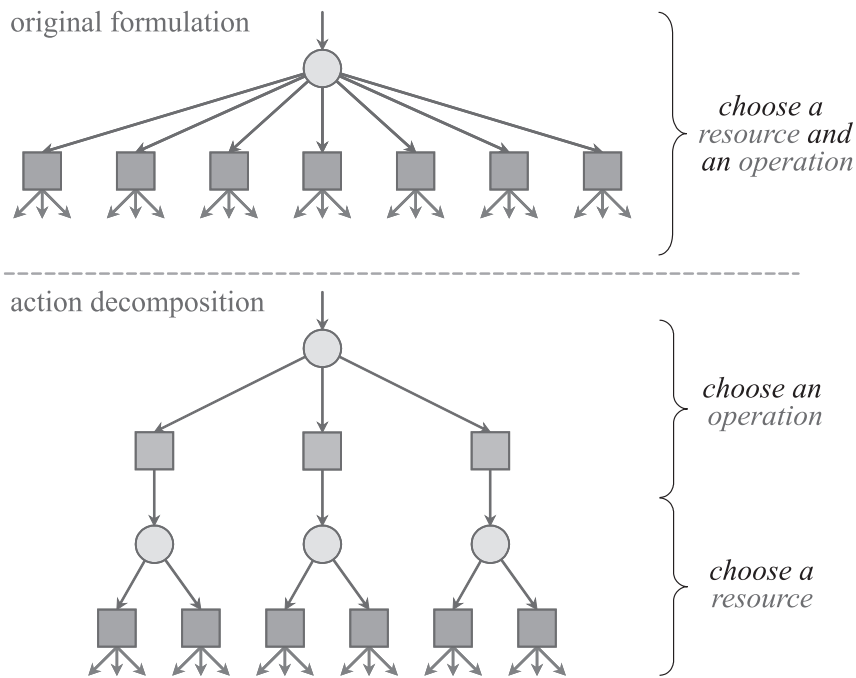

Figure 7: Action selection before (up) and after (down) action space decomposition.

### 3.3.3 Clustering the Tasks

The idea of divide-and-conquer is widely used in artificial intelligence and recently it has appeared in the theory of dealing with large-scale MDPs. Partitioning a problem into several smaller subproblems is also often applied to decrease computational complexity in combinatorial optimization problems, for example, in scheduling theory.

We propose a simple and still efficient partitioning method for a practically very important class of performance measures. In real world situations the tasks very often have release dates and due dates, and the performance measure, e.g., total lateness and number of tardy tasks, depends on meeting the deadlines. Note that these measures are regular. We denote the (possibly randomized) functions defining the *release* and *due dates* of the tasks by $A : \mathcal{T} \to \mathbb{N}$ and $B : \mathcal{T} \to \mathbb{N}$, respectively. In this section we restrict ourselves to performance measures that are regular and depend on due dates. In order to cluster the tasks, we need the definition of *weighted expected slack time* which is given as follows

$$S_w(v) = \sum_{s \in \Gamma(v)} w(s) \, \mathbb{E}\Big[ B(v) - A(v) - D(s, v) \Big],$$

where $\Gamma(v) = \{ s \in \mathcal{S} \mid \langle s, v \rangle \in dom(D) \}$ denotes the set of resource states in which task $v$ can be processed, and $w(s)$ are weights corresponding, for example, to the likelihood that resource state $s$ appears during execution, or they can be simply $w(s) = 1 / |\Gamma(v)|$.

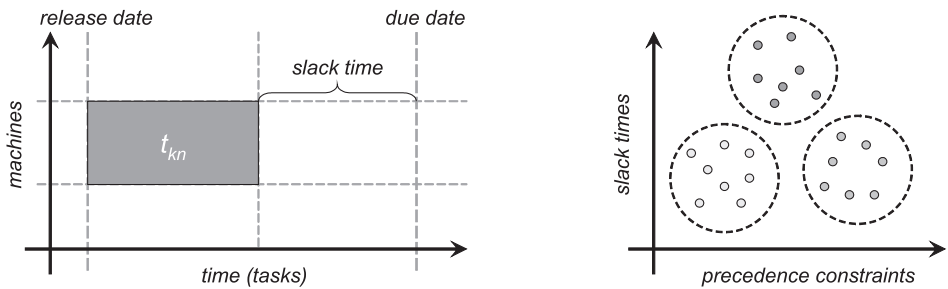

Figure 8: Clustering the tasks according to their slack times and precedence constraints.

In order to increase computational speed, we suggest clustering the tasks in $\mathcal{T}$ into successive disjoint subsets $\mathcal{T}_1, \ldots, \mathcal{T}_k$ according to the precedence constraints and the expected slack times; take a look at Figure 8 for an illustration. The basic idea behind our approach is that we should handle the most constrained tasks first. Therefore, *ideally*, if $\mathcal{T}_i$ and $\mathcal{T}_j$ are two clusters and $i < j$, then tasks in $\mathcal{T}_i$ had expected slack times smaller than tasks in $\mathcal{T}_j$. However, in most of the cases clustering is not so simple, since the precedence constraints must also be taken into account and this clustering criterion has the priority. Thus, if $\langle u, v \rangle \in \mathcal{C}$, $u \in \mathcal{T}_i$ and $v \in \mathcal{T}_j$, then $i \leq j$ must hold. During learning, first, tasks in $\mathcal{T}_1$ are allocated to resources, only. After some episodes, we fix the allocation policy concerning tasks in $\mathcal{T}_1$ and we start sampling to achieve a good policy for tasks in $\mathcal{T}_2$, and so on.

Naturally, clustering the tasks is a two-edged weapon, making too small clusters may seriously decrease the performance of the best achievable policy, making too large clusters may considerably slow down the system. This technique, however, has several advantages, e.g., (1) it effectively decreases the search space; (2) further reduces the number of available actions in the states; and, additionally (3) speeds up the learning, since the sample trajectories become smaller (only a small part of the tasks is allocated in a trial and, consequently, the variance of the total costs is decreased). The effects of clustering relative to the size of the clusters were analyzed experimentally and are presented in Section 4.5.

### 3.3.4 DISTRIBUTED SAMPLING

Finally, we argue that the presented approach can be easily modified in order to allow computing a policy on several processors in a distributed way. Parallel computing can further speed up the calculation of the solution. We will consider extensions of the algorithm using both *shared memory* and *distributed memory* architectures. Let us suppose we have $k$ processors, and denote the set of all processors by $\mathcal{P} = \{p_1, p_2, \ldots, p_k\}$.

In case we have a parallel system with a shared memory architecture, e.g., UMA (uniform memory access), then it is straightforward to parallelize the computation of a control policy. Namely, each processor $p \in \mathcal{P}$ can sample the search space independently, while by using the same, shared value function. The (joint) policy can be calculated using this common, global value function, e.g., the greedy policy w.r.t. this function can be applied.

Parallelizing the solution by using an architecture with distributed memory is more challenging. Probably the simplest way to parallelize our approach to several processors with distributed memory is to let the processors search independently by letting them working with their own, local value functions. After a given time or number of iterations, we may treat the best achieved solution as the joint policy. More precisely, if we denote the aggregated initial state by $x_0$, then the joint control policy $\pi$ can be defined as follows

$$\pi \in \argmin_{\pi_p\,(p \in \mathcal{P})} J^{\pi_p}(x_0) \qquad \text{or} \qquad \pi \in \argmin_{\pi_p\,(p \in \mathcal{P})} \min_{a \in \mathcal{A}(x_0)} Q^{\pi_p}(x_0, a),$$

where $J^{\pi_p}$ and $Q^{\pi_p}$ are (approximate) state- and action-value functions calculated by processor $p \in \mathcal{P}$. Control policy $\pi_p$ is the solution of processor $p$ after a given number of iterations. During our numerical experiments we usually applied $10^4$ iterations.

Naturally, there could be many (more sophisticated) ways to parallelize the computation using several processors with distributed memory. For example, from time to time the processors could exchange some of their best episodes (trajectories with the lowest costs) and learn from the experiments of the others. In this way, they could help improve the value functions of each other. Our numerical experiments, presented in Section 4.3, showed that even in the simplest case, distributing the calculation speeds up the optimization considerably. Moreover, in the case of shared memory the speedup was almost linear.

As parallel computing represents a very promising way do deal with large-scale systems, their further theoretical and experimental investigation would be very important. For example, by harmonizing the exploration of the processors, the speedup could be improved.

## 4. Experimental Results

In this section some experimental results on both benchmark and industry-related problems are presented. These experiments highlight some characteristics of the solution.

## 4.1 Testing Methodology

In order to experimentally study our resource control approach, a simulation environment was developed in C++. We applied FQL and, in most of the cases, SVRs which were realized by the LIBSVM free library for support vector machines (Chang & Lin, 2001). After centering and scaling the data into interval $[0, 1]$, we used Gaussian kernels and shrinking techniques. We always applied rollout algorithms and action decomposition, but clustering was only used in tests presented in Section 4.5, furthermore, distributed sampling was only applied in test shown in Section 4.3. In both of the latter cases (tests for clustering and distributed sampling) we used hash tables with approximately 256Mb hash memory.

The performance of the algorithm was measured as follows. Testing took place in two main fields: the first one was a benchmark scheduling dataset of hard problems, the other one was a simulation of a "real world" production control problem. In the first case the best solution, viz., the optimal value of the (aggregated) initial state, $J^*(x_0) = \min_a Q^*(x_0, a)$, was known for most of the test instances. Some "very hard" instances occurred for which only lower and upper bounds were known, e.g., $J_1^*(x_0) \leq J^*(x_0) \leq J_2^*(x_0)$. In these cases we assumed that $J^*(x_0) \approx (J_1^*(x_0) + J_2^*(x_0))/2$. Since these estimations were "good" (viz., the length of the intervals were short), this simplification might not introduce considerable error to our performance estimations. In the latter test case we have generated the problems with a generator in a way that $J^*(x_0)$ was known concerning the constructed problems.

The performance presented in the tables of the section, more precisely the average, $\overline{E}_i$, and the standard deviation, $\sigma(E_i)$, of the error in iteration $i$ were computed as follows

$$\overline{E}_i = \frac{1}{N} \sum_{j=1}^{N} \left[ G_j^i - J^*(x_0) \right], \qquad \text{and} \qquad \sigma(E_i) = \sqrt{\frac{1}{N} \sum_{j=1}^{N} \left[ G_j^i - J^*(x_0) - \overline{E}_i \right]^2},$$

where $G_j^i$ denotes the cumulative incurred costs in iteration $i$ of sample $j$ and $N$ is the sample size. Unless indicated otherwise, the sample contained the results of 100 simulation trials for each parameter configuration (which is associated with the rows of the tables).

As it was shown in Section 2.5.2, RAP-MDPs are aperiodic, moreover, they have the APP property, therefore, discounting is not necessary to achieve a well-defined problem. However, in order to enhance learning, it is still advised to apply discounting and, therefore, to give less credit to events which are farther from the current decision point. Heuristically, we suggest applying $\alpha = 0.95$ for middle-sized RAPs (e.g., with few hundreds of tasks), such as the problems of the benchmark dataset, and $\alpha = 0.99$ for large-scale RAPs (e.g., with few thousands of tasks), such as the problems of the industry-related experiments.

## 4.2 Benchmark Datasets

The ADP based resource control approach was tested on Hurink's benchmark dataset (Hurink, Jurisch, & Thole, 1994). It contains *flexible job-shop scheduling problems* (FJSPs) with 6–30 jobs (30–225 tasks) and 5–15 machines. The applied performance measure is the maximum completion time of the tasks (makespan). These problems are "hard", which means, e.g., that standard dispatching rules or heuristics perform poorly on them. This dataset consists of four subsets, each subset contains about 60 problems. The subsets (sdata, edata, rdata, vdata) differ in the ratio of machine interchangeability (flexibility), which is shown in the "flex(ib)" columns in Tables 3 and 2. The columns with label "n iters" (and "avg err") show the average error after carrying out altogether "n" iterations. The "std dev" columns in the tables of the section contain the standard deviation of the sample.

Table 2 illustrates the performance on some typical dataset instances and also gives some details on them, e.g., the number of machines and jobs (columns with labels "mcs" and "jbs"). In Table 3 the summarized performance on the benchmark datasets is shown.

| benchmark configuration | | | | | average error (standard deviation) | | |
|---|---|---|---|---|---|---|---|
| dataset | inst | mcs | jbs | flex | 1000 iters | 5000 iters | 10 000 iters |
| sdata | mt06 | 6 | 6 | 1 | 1.79 (1.01) % | 0.00 (0.00) % | 0.00 (0.00) % |
| sdata | mt10 | 10 | 10 | 1 | 9.63 (4.59) % | 8.83 (4.37) % | 7.92 (4.05) % |
| sdata | la09 | 5 | 15 | 1 | 5.67 (2.41) % | 3.87 (1.97) % | 3.05 (1.69) % |
| sdata | la19 | 10 | 10 | 1 | 11.65 (5.21) % | 6.44 (3.41) % | 3.11 (1.74) % |
| sdata | la39 | 15 | 15 | 1 | 14.61 (7.61) % | 12.74 (5.92) % | 11.92 (5.63) % |
| sdata | la40 | 15 | 15 | 1 | 10.98 (5.04) % | 8.87 (4.75) % | 8.39 (4.33) % |
| edata | mt06 | 6 | 6 | 1.15 | 0.00 (0.00) % | 0.00 (0.00) % | 0.00 (0.00) % |
| edata | mt10 | 10 | 10 | 1.15 | 18.14 (8.15) % | 12.51 (6.12) % | 9.61 (4.67) % |
| edata | la09 | 5 | 15 | 1.15 | 7.51 (3.33) % | 5.23 (2.65) % | 2.73 (1.89) % |
| edata | la19 | 10 | 10 | 1.15 | 8.04 (4.64) % | 4.14 (2.81) % | 1.38 (1.02) % |
| edata | la39 | 15 | 15 | 1.15 | 22.80 (9.67) % | 17.32 (8.29) % | 12.41 (6.54) % |
| edata | la40 | 15 | 15 | 1.15 | 14.78 (7.14) % | 8.08 (4.16) % | 6.68 (4.01) % |
| rdata | mt06 | 6 | 6 | 2 | 6.03 (3.11) % | 0.00 (0.00) % | 0.00 (0.00) % |
| rdata | mt10 | 10 | 10 | 2 | 17.21 (8.21) % | 12.68 (6.81) % | 7.87 (4.21) % |
| rdata | la09 | 5 | 15 | 2 | 7.08 (3.23) % | 6.15 (2.92) % | 3.80 (2.17) % |
| rdata | la19 | 10 | 10 | 2 | 18.03 (8.78) % | 11.71 (5.78) % | 8.87 (4.38) % |
| rdata | la39 | 15 | 15 | 2 | 24.55 (9.59) % | 18.90 (8.05) % | 13.06 (7.14) % |
| rdata | la40 | 15 | 15 | 2 | 23.90 (7.21) % | 18.91 (6.92) % | 14.08 (6.68) % |
| vdata | mt06 | 6 | 6 | 3 | 0.00 (0.00) % | 0.00 (0.00) % | 0.00 (0.00) % |
| vdata | mt10 | 10 | 10 | 5 | 8.76 (4.65) % | 4.73 (2.23) % | 0.45 (0.34) % |
| vdata | la09 | 5 | 15 | 2.5 | 9.92 (5.32) % | 7.97 (3.54) % | 4.92 (2.60) % |
| vdata | la19 | 10 | 10 | 5 | 14.42 (7.12) % | 11.61 (5.76) % | 6.54 (3.14) % |
| vdata | la39 | 15 | 15 | 7.5 | 16.16 (7.72) % | 12.25 (6.08) % | 9.02 (4.48) % |
| vdata | la40 | 15 | 15 | 7.5 | 5.86 (3.11) % | 4.08 (2.12) % | 2.43 (1.83) % |

Table 2: Performance (average error and deviation) on some typical benchmark problems.

Simple dispatching rules (which are often applied in practice), such as greedy ones, perform poorly on these benchmark datasets. Their average error is around 25–30 %. In contrast, Table 3 demonstrates that using our method, the average error is less than 5 % after 10 000 iterations. It shows that learning is beneficial for this type of problems.

The best performance on these benchmark datasets was achieved by Mastrolilli and Gambardella (2000). Though, their algorithm performs slightly better than ours, their solution exploits the (unrealistic) specialties of the dataset, e.g., the durations do not depend on the resources; the tasks are linearly ordered in the jobs; each job consists of the same number of tasks. Moreover, it cannot be easily generalized to stochastic resource control problem our algorithm faces. Therefore, the comparison of the solutions is hard.

| benchmark | | 1000 iterations | | 5000 iterations | | 10 000 iterations | |
|---|---|---|---|---|---|---|---|
| dataset | flexib | avg err | std dev | avg err | std dev | avg err | std dev |
| sdata | 1.0 | 8.54 % | 5.02 % | 5.69 % | 4.61 % | 3.57 % | 4.43 % |
| edata | 1.2 | 12.37 % | 8.26 % | 8.03 % | 6.12 % | 5.26 % | 4.92 % |
| rdata | 2.0 | 16.14 % | 7.98 % | 11.41 % | 7.37 % | 7.14 % | 5.38 % |
| vdata | 5.0 | 10.18 % | 5.91 % | 7.73 % | 4.73 % | 3.49 % | 3.56 % |
| average | 2.3 | 11.81 % | 6.79 % | 8.21 % | 5.70 % | 4.86 % | 4.57 % |

Table 3: Summarized performance (average error and deviation) on benchmark datasets.

## 4.3 Distributed Sampling

The possible parallelizations of the presented method was also investigated, i.e., the *speedup* of the system relative to the number of processors (in practise, the multiprocessor environment was emulated on a single processor, only). The average number of iterations was studied, until the system could reach a solution with less than 5 % error on Hurink's dataset. The average speed of a single processor was treated as a unit, for comparison.

In Figure 9 two cases are shown: in the first case (rear dark bars) each processor could access a common global value function. It means that each processor could read and write the same global value function, but otherwise, they searched (sampled the search space) independently. Figure 9 demonstrates that in this case the speedup was almost linear.

In the second case (front light bars) each processor had its own (local) value function (which is more realistic in a strongly distributed system, such as a GRID) and, after the search had been finished, these individual value functions were compared. Therefore, all of the processors had estimations of their own, and after the search, the local solution of the best performing processor was selected. Figure 9 shows the achieved speedup in case we stopped the simulation if any of the processors achieved a solution with less than 5 % error.

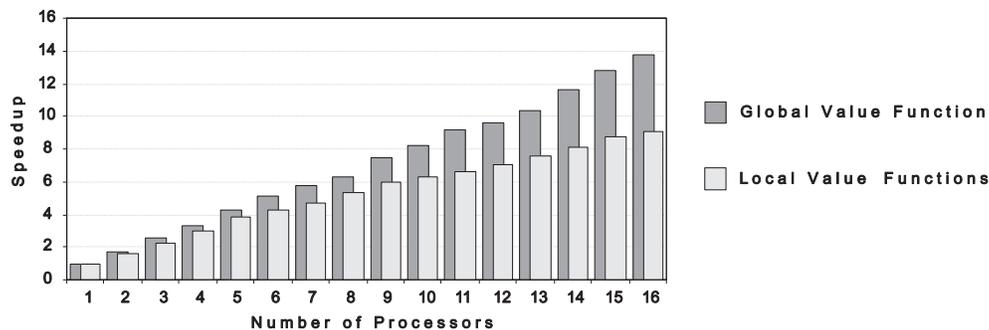

Figure 9: Average speedup relative to the number of processors.

The experiments show that the computation of the resource allocator function can be effectively distributed, even if there is not a commonly accessible value function available.

## 4.4 Industry Related Tests

We also initiated experiments on a simulated factory by modeling the structure of a real plant producing customized mass-products, especially, light bulbs. These industrial data came from a huge national industry-academia project, for research and development of solutions which support manufacturing enterprises in coping with the requirements of adaptiveness, realtimeness and cooperativeness (Monostori, Kis, Kádár, Váncza, & Erdős, 2008).

| optimal slack ratio | 1000 iterations | | 5000 iterations | | 10 000 iterations | |
|---|---|---|---|---|---|---|
| | avg err | std dev | avg err | std dev | avg err | std dev |
| 50 % | 0.00 % | 0.00 % | 0.00 % | 0.00 % | 0.00 % | 0.00 % |
| 40 % | 0.12 % | 0.10 % | 0.00 % | 0.00 % | 0.00 % | 0.00 % |
| 30 % | 0.52 % | 0.71 % | 0.24 % | 0.52 % | 0.13 % | 0.47 % |
| 20 % | 1.43 % | 1.67 % | 1.11 % | 1.58 % | 1.05 % | 1.49 % |
| 10 % | 5.28 % | 3.81 % | 4.13 % | 3.53 % | 3.91 % | 3.48 % |
| 0 % | 8.89 % | 5.17 % | 7.56 % | 5.04 % | 6.74 % | 4.83 % |

Table 4: Summarized performance relative to the optimal slack ratio of the system.

Since, we did not have access to historical data concerning past orders, we used randomly generated orders (jobs) with random due dates. The tasks and the process-plans of the jobs, however, covered real products; as well as, the resources covered real machine types. In this plant the machines require product-type dependent setup times, and there are some special tasks that have durations but that do not require any resources to be processed, for example, cooling down. Another feature of the plant is that at some previously given time points preemptions are allowed, e.g., at the end of a work shift. The applied performance measure was to minimize the *number of late jobs*, viz., jobs that are finished after their due dates, and an additional secondary measure was to minimize the total cumulative *lateness*, which can be applied to compare two schedules having the same number of late jobs.

During these experiments the jobs and their due dates were generated by a special parameterizable generator in a way that optimally none of the jobs were late. Therefore, it was known that $J^*(x_0) = 0$ and the error of the algorithm was computed accordingly.

In the first case, shown in Table 4, we applied 16 machines and 100 random jobs, which altogether contained more than 200 tasks. The convergence properties were studied relative to the optimal *slack ratio*. In the deterministic case, e.g., the slack ratio of a solution is

$$\Phi(\varrho) = \frac{1}{n} \sum_{i=1}^{n} \frac{B(J_i) - F(J_i)}{B(J_i) - A(J_i)},$$

where $n$ is the number of jobs; $A(J)$ and $B(J)$ denote the release and due date of job $J$, respectively; $F(J)$ is the finish time of job $J$ relative to solution $\varrho$, namely, the latest finish time of the tasks in the job. Roughly, the slack ratio measures the tightness of the solution, for example, if $\Phi(\varrho) > 0$, then it shows that the jobs were, on the average, finished before their due dates and if $\Phi(\varrho) < 0$, then it indicates that, approximately, many jobs were late. If $\Phi(\varrho) = 0$, then it shows that if all the jobs meet their due dates, each job was finished just in time, there were no spare (residual) times. Under the *optimal* slack ratio we mean the maximal achievable slack ratio (by an optimal solution). During the experiments these values were known because of the special construction of the test problem instances. We applied the optimal slack ratio to measure how "hard" a problem is. The first column of Table 4 shows the optimal slack ratio in percentage, e.g., 30 % means a 0.3 slack ratio.

| configuration | | 1000 iterations | | 5000 iterations | | 10 000 iterations | |
|---|---|---|---|---|---|---|---|
| machs | tasks | avg err | std dev | avg err | std dev | avg err | std dev |
| 6 | 30 | 4.01 % | 2.24 % | 3.03 % | 1.92 % | 2.12 % | 1.85 % |
| 16 | 140 | 4.26 % | 2.32 % | 3.28 % | 2.12 % | 2.45 % | 1.98 % |
| 25 | 280 | 7.05 % | 2.55 % | 4.14 % | 2.16 % | 3.61 % | 2.06 % |
| 30 | 560 | 7.56 % | 3.56 % | 5.96 % | 2.47 % | 4.57 % | 2.12 % |
| 50 | 2000 | 8.69 % | 7.11 % | 7.24 % | 5.08 % | 6.04 % | 4.53 % |
| 100 | 10000 | 15.07 % | 11.89 % | 10.31% | 7.97 % | 9.11 % | 7.58 % |

Table 5: Summarized performance relative to the number of machines and tasks.

In the second case, shown in Table 5, we have fixed the optimal slack ratio of the system to 10 % and investigated the convergence speed relative to the plant size (number of machines) and the number of tasks. In the last two experiments (configuration having 2000 and 10 000 tasks) only 10 samples were generated, because of the long runtime. The computation of 10 000 iterations took approximately 30 minutes for the 50 machines & 2000 tasks configuration and 3 hours for the 100 machines & 10000 tasks configuration[1].

The results demonstrate that the ADP and adaptive sampling based solution scales well with both the slack ratio and the size (the number of machines and task) of the problem.

## 4.5 Clustering Experiments

The effectiveness of clustering on industry-related data was also studied. We considered a system with 60 resources and 1000 random tasks distributed among 400–500 jobs (there were approximately 1000–2000 precedence constraints). The tasks were generated in a way that, in the optimal case, none of them are late and the slack ratio is about 20 %.

First, the tasks were ordered according to their slack times and then they were clustered. We applied $10^4$ iterations on each cluster. The computational time in case of using only one cluster was treated as a unit. In Table 6 the average and the standard deviation of the error and the computational speedup are shown relative to the number tasks in a cluster.

The results demonstrate that partitioning the search space not only results in a greater speed, but it is often accompanied by better solutions. The latter phenomenon can be explained by the fact that using smaller sample trajectories generates smaller variance that is preferable for learning. On the other hand, making too small clusters may decrease the performance (e.g., making 50 clusters with 20 tasks in the current case). In our particular case applying 20 clusters with approximately 50 tasks in each cluster balances good performance (3.02 % error on the average) with remarkable speedup (approximately 3.28 ×).

---

1. The tests were performed on a Centrino (Core-Duo) 1660Mhz CPU ($\approx$ P4 3GHz) with 1Gb RAM.

| configuration | | performance after 10 000 iterations per cluster | | | | |
|---|---|---|---|---|---|---|
| clusters | tasks | late jobs | avg error | std dev | speed | speedup |
| 1 | 1000 | 28.1 | 6.88 % | 2.38 % | 423 s | 1.00 × |
| 5 | 200 | 22.7 | 5.95 % | 2.05 % | 275 s | 1.54 × |
| 10 | 100 | 20.3 | 4.13 % | 1.61 % | 189 s | 2.24 × |
| 20 | 50 | 13.9 | 3.02 % | 1.54 % | 104 s | 3.28 × |
| 30 | 33 | 14.4 | 3.15 % | 1.51 % | 67 s | 6.31 × |
| 40 | 25 | 16.2 | 3.61 % | 1.45 % | 49 s | 8.63 × |
| 50 | 20 | 18.7 | 4.03 % | 1.43 % | 36 s | 11.65 × |

Table 6: Speedup and performance relative to the number of tasks in a cluster.

As clustering the tasks represents a considerable help in dealing with large-scale RAPs, their further theoretical and experimental investigations are very promising.

## 4.6 Adaptation to Disturbances

In order to verify the proposed algorithm in changing environments, experiments were initiated and carried out on random JSPs with the aim of minimizing the makespan. The adaptive features of the system were tested by confronting it with unexpected events, such as: resource breakdowns, new resource availability (Figure 10), new job arrivals or job cancellations (Figure 11). In Figures 10 and 11 the horizontal axis represents time, while the vertical one, the achieved performance measure. The figures were made by averaging hundred random samples. In these tests 20 machines were used with few dozens of jobs.

During each test episode there was an unexpected event (disturbance) at time $t = 100$. After the change took place, we considered two possibilities: we either restarted the iterative scheduling process from scratch or continued the learning, using the current (obsolete) value function. We experienced that the latter approach is much more efficient.

An explanation of this phenomenon can be that the value functions of control policies and the optimal value function Lipschitz continuously depend on the transition-probability and the immediate-cost functions of the MDP (Csáji, 2008). Therefore, small changes in the environmental dynamics cannot cause arbitrary large changes in the value function. The results of our numerical experiments, shown in Figures 10 and 11, are indicative of the phenomenon that the average change of the value function is not very large. Consequently, applying the obsolete value function after a change took place in the MDP is preferable over restarting the whole optimization process from scratch. This adaptive feature makes ADP/RL based approaches even more attractive for practical applications.

The results, black curves, show the case when the obsolete value function approximation was applied after the change took place. The performance which would arise if the system recomputed the whole schedule from scratch is drawn in gray in part (a) of Figure 10.

One can notice that even if the problem became "easier" after the change in the environment (at time $t = 100$), for example, a new resource was available (part (b) of Figure 10) or a job was cancelled (part (b) of Figure 11), the performance started to slightly decrease ($\kappa$ started to slightly increase) after the event. This phenomenon can be explained by the fact that even in these special cases the system had to "explore" the new configuration.

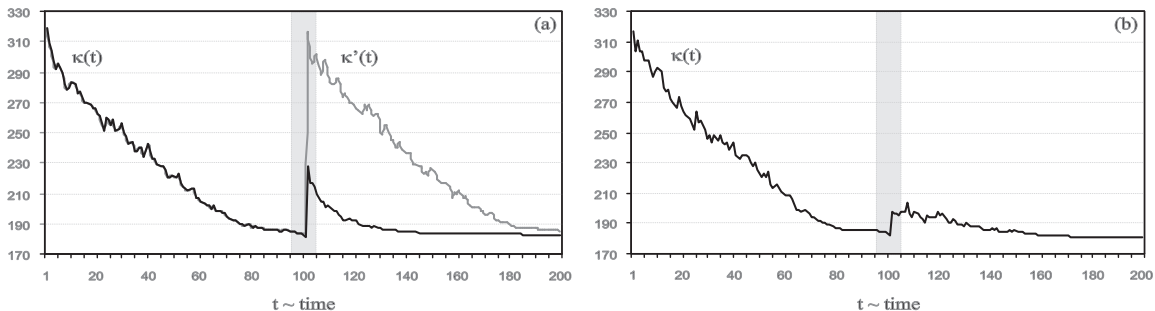

Figure 10: The black curves, $\kappa(t)$, show the performance measure in case there was a resource breakdown (a) or a new resource availability (b) at $t = 100$; the gray curve, $\kappa'(t)$, demonstrates the case the policy would be recomputed from scratch.

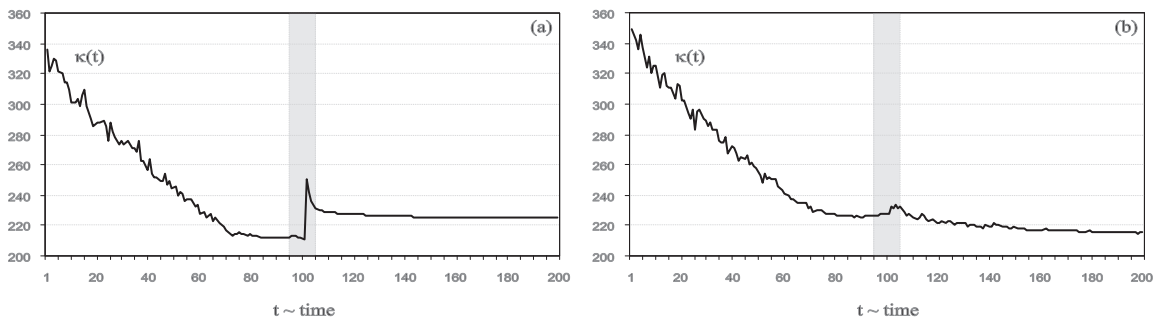

Figure 11: The black curves, $\kappa(t)$, show the performance measure during resource control in case there was a new job arrival (a) or a job cancellation (b) at time $t = 100$.

## 5. Concluding Remarks

Efficient allocation of scarce, reusable resources over time in uncertain and dynamic environments is an important problem that arises in many real world domains, such as production control. The paper took a machine learning (ML) approach to this problem. First, a general resource allocation framework was presented and, in order to define reactive solutions, it was reformulated as a stochastic shortest path problem, a special Markov decision process (MDP). The core idea of the solution was the application of approximate dynamic programming (ADP) and reinforcement learning (RL) techniques with Monte Carlo simulation to stochastic resource allocation problems (RAPs). Regarding compact value function representations, two approaches were studied: hash table and support vector regression (SVR), specially $\nu$-SVRs. Afterwards, several additional improvements, such as the application of limited-lookahead rollout algorithms in the initial phases, action space decomposition, task clustering and distributed sampling, were suggested for speeding up the computation of a good control policy. Finally, the effectiveness of the approach was demonstrated by results of numerical simulation experiments on both benchmark and industry-related data. These experiments also supported the adaptive capabilities of the proposed method.

There are several advantages why ML based resource allocation is preferable to other kinds of RAP solutions, e.g., classical approaches. These favorable features are as follows:

1. The presented RAP framework is very *general*, it can model several resource management problems that appear in practice, such as scheduling problems, transportation problems, inventory management problems or maintenance and repair problems.

2. ADP/RL based methods essentially face the problem under the presence of *uncertainties*, since their theoretical foundation is provided by MDPs. Moreover, they can adapt to unexpected *changes* in the environmental dynamics, such as breakdowns.

3. Additionally, for most algorithms theoretical *guarantees* of finding (approximately) optimal solutions, at least in the limit, are known. As demonstrated by our experiments, the actual *convergence speed* for RAPs is usually high, especially in the case of applying the described improvements, such as clustering or distributed sampling.

4. The simulation experiments on industrial data also demonstrate that ADP/RL based solutions *scale well* with the workload and the size of the problem and, therefore, they can be effectively applied to handle real world RAPs, such as production scheduling.

5. *Domain specific knowledge* can also be incorporated into the solution. The base policy of the rollout algorithm, for example, can reflect a priori knowledge about the structure of the problem; later this knowledge may appear in the exploration strategy.

6. Finally, the proposed method constitutes an *any-time* solution, since the sampling can be stopped after any number of iterations. By this way, the amount of computational time can be controlled, which is also an important practical advantage.

Consequently, ML approaches have great potentials in dealing with real world RAPs, since they can handle large-scale problems even in dynamic and uncertain environments.

Several further research directions are possible. Now, as a conclusion to the paper, we highlight some of them. The suggested improvements, such as *clustering* and *distributed sampling*, should be further investigated since they resulted in considerable speedup. The guidance of reinforcement learning with *rollout* algorithms might be effectively applied in other applications, as well. The theoretical analysis of the *average* effects of environmental changes on the value functions could result in new approaches to handle disturbances. Another promising direction would be to extend the solution in a way which also takes *risk* into account and, e.g., minimizes not only the expected value of the total costs but also the *deviation*, as a secondary criterion. Finally, trying to apply the solution in a *pilot* project to control a real plant would be interesting and could motivate further research directions.

## 6. Acknowledgments

The work was supported by the Hungarian Scientific Research Fund (OTKA), Grant No. T73376, and by the EU-funded project "Coll-Plexity", Grant No. 12781 (NEST). Balázs Csanád Csáji greatly acknowledges the scholarship of the Hungarian Academy of Sciences. The authors express their thanks to Tamás Kis for his contribution related to the tests on industrial data and to Csaba Szepesvári for the helpful discussions on machine learning.

# References

Andrieu, C., Freitas, N. D., Doucet, A., & Jordan, M. I. (2003). An introduction to MCMC (Markov Chain Monte Carlo) for machine learning. *Machine Learning, 50*, 5–43.

Aydin, M. E., & Öztemel, E. (2000). Dynamic job-shop scheduling using reinforcement learning agents. *Robotics and Autonomous Systems, 33*, 169–178.

Beck, J. C., & Wilson, N. (2007). Proactive algorithms for job shop scheduling with probabilistic durations. *Journal of Artificial Intelligence Research, 28*, 183–232.

Bellman, R. E. (1961). *Adaptive Control Processes*. Princeton University Press.

Bertsekas, D. P. (2005). Dynamic programming and suboptimal control: A survey from ADP to MPC. *European Journal of Control, 11*(4–5), 310–334.

Bertsekas, D. P., & Tsitsiklis, J. N. (1996). *Neuro-Dynamic Programming*. Athena Scientific, Belmont, Massachusetts.

Bertsekas, D. P. (2001). *Dynamic Programming and Optimal Control* (2nd edition). Athena Scientific, Belmont, Massachusetts.

Bulitko, V., & Lee, G. (2006). Learning in real-time search: A unifying framework. *Journal of Artificial Intelligence Research, 25*, 119–157.

Chang, C. C., & Lin, C. J. (2001). LIBSVM: A library for support vector machines. Software available on-line at http://www.csie.ntu.edu.tw/~cjlin/libsvm.

Csáji, B. Cs. (2008). *Adaptive Resource Control: Machine Learning Approaches to Resource Allocation in Uncertain and Changing Environments*. Ph.D. thesis, Faculty of Informatics, Eötvös Loránd University, Budapest.

Csáji, B. Cs., Kádár, B., & Monostori, L. (2003). Improving multi-agent based scheduling by neurodynamic programming. In *Proceedings of the 1st International Conference on Holonic and Mult-Agent Systems for Manufacturing, September 1–3, Prague, Czech Republic*, Vol. 2744 of *Lecture Notes in Artificial Intelligence*, pp. 110–123.

Csáji, B. Cs., & Monostori, L. (2006). Adaptive sampling based large-scale stochastic resource control. In *Proceedings of the 21st National Conference on Artificial Intelligence (AAAI 2006), July 16–20, Boston, Massachusetts*, pp. 815–820.

Dolgov, D. A., & Durfee, E. H. (2006). Resource allocation among agents with MDP-induced preferences. *Journal of Artificial Intelligence Research, 27*, 505–549.

Even-Dar, E., & Mansour, Y. (2003). Learning rates for Q-learning. *Journal of Machine Learning Research, 5*, 1–25.

Feinberg, E. A., & Shwartz, A. (Eds.). (2002). *Handbook of Markov Decision Processes: Methods and Applications*. Kluwer Academic Publishers.

Gersmann, K., & Hammer, B. (2005). Improving iterative repair strategies for scheduling with the SVM. *Neurocomputing, 63*, 271–292.

Hastings, W. K. (1970). Monte Carlo sampling methods using Markov chains and their application. *Biometrika, 57*, 97–109.

Hatvany, J., & Nemes, L. (1978). Intelligent manufacturing systems - a tentative forecast. In Niemi, A. (Ed.), *A link between science and applications of automatic control; Proceedings of the 7th IFAC World Congress*, Vol. 2, pp. 895–899.

Hurink, E., Jurisch, B., & Thole, M. (1994). Tabu search for the job shop scheduling problem with multi-purpose machines. *Operations Research Spektrum*, *15*, 205–215.

Kirkpatrick, S., Gelatt, C. D., & Vecchi, M. P. (1983). Optimization by simulated annealing. *Science*, *220*(4598), 671–680.

Mastrolilli, M., & Gambardella, L. M. (2000). Effective neighborhood functions for the flexible job shop problem. *Journal of Scheduling*, *3*(1), 3–20.

Metropolis, N., Rosenbluth, A., Rosenbluth, M., Teller, A., & Teller, E. (1953). Equation of state calculations by fast computing machines. *Journal of Chemical Physics*, *21*, 1087–1092.

Monostori, L., Kis, T., Kádár, B., Váncza, J., & Erdős, G. (2008). Real-time cooperative enterprises for mass-customized production. *International Journal of Computer Integrated Manufacturing*, (to appear).

Papadimitriou, C. H. (1994). *Computational Complexity*. Addison-Wesley.

Pinedo, M. (2002). *Scheduling: Theory, Algorithms, and Systems*. Prentice-Hall.

Powell, W. B., & Van Roy, B. (2004). *Handbook of Learning and Approximate Dynamic Programming*, chap. Approximate Dynamic Programming for High-Dimensional Resource Allocation Problems, pp. 261–283. IEEE Press, Wiley-Interscience.

Riedmiller, S., & Riedmiller, M. (1999). A neural reinforcement learning approach to learn local dispatching policies in production scheduling. In *Proceedings of the 16th International Joint Conference on Artificial Intelligence, Stockholm, Sweden*, pp. 764–771.

Schneider, J. G., Boyan, J. A., & Moore, A. W. (1998). Value function based production scheduling. In *Proceedings of the 15th International Conference on Machine Learning*, pp. 522–530. Morgan Kaufmann, San Francisco, California.

Schölkopf, B., Smola, A., Williamson, R. C., & Bartlett, P. L. (2000). New support vector algorithms. *Neural Computation*, *12*, 1207–1245.

Singh, S., Jaakkola, T., Littman, M., & Szepesvári, Cs. (2000). Convergence results for single-step on-policy reinforcement-learning algorithms. *Machine Learning*, *38*(3), 287–308.

Sontag, E. D. (1998). *Mathematical Control Theory: Deterministic Finite Dimensional Systems*. Springer, New York.

Sutton, R. S., & Barto, A. G. (1998). *Reinforcement learning*. The MIT Press.

Topaloglu, H., & Powell, W. B. (2005). A distributed decision-making structure for dynamic resource allocation using nonlinear function approximators. *Operations Research*, *53*(2), 281–297.

Zhang, W., & Dietterich, T. (1995). A reinforcement learning approach to job-shop scheduling. In *Proceedings of the 14th International Joint Conference on Artificial Intelligence*, pp. 1114–1120. Morgan Kauffman.